\documentclass[10pt,conference,a4paper]{IEEEtran}
\makeatletter
\newcommand{\RNum}[1]{\uppercase\expandafter{\romannumeral #1\relax}}
\newcommand{\tabincell}[2]{\begin{tabular}{@{}#1@{}}#2\end{tabular}} 
\makeatother
\usepackage{geometry}
\usepackage{amsfonts}
\usepackage{graphicx}
\usepackage{amsmath}
\usepackage{subfig}
\usepackage{multirow}
\usepackage{booktabs}
\usepackage{algorithm}
\usepackage{algorithmic}
\usepackage{CJK}
\usepackage{color}
\usepackage{amsthm}
\usepackage{setspace}
\usepackage{bm}
\usepackage{caption}
\usepackage{verbatim}
\usepackage{enumerate}
\usepackage{cite}

\usepackage{float}
\makeatletter
\newcommand*{\rom}[1]{\expandafter\@slowromancap\romannumeral #1@}
\makeatother
\begin{document}
\newtheorem{myDef}{Definition}

\begin{comment}
$^1$School of Information Science and Engineering, Southeast University, Nanjing 210096, China\\
$^2$School of automation, Southeast University, Nanjing 210096, China\\
\end{comment}

\title{An Attention-Based Approach for Single Image Super Resolution}
\author{\IEEEauthorblockN{Yuan Liu$^{1,2,3}$, Yuancheng Wang$^{1}$, Nan Li$^{1}$,Xu Cheng$^{4}$, Yifeng Zhang$^{1,2,3,*}$, Yongming Huang$^{1}$, Guojun Lu$^{5}$}
\IEEEauthorblockA{$^1$School of Information Science and Engineering, Southeast University, Nanjing 210096, China\\
$^2$Nanjing Institute of Communications Technologies, Nanjing, 211100, China\\
$^3$State Key Lab. for Novel Software Technology, Nanjing University, Nanjing, 210093, China\\
$^4$School of automation, Southeast University, Nanjing 210096, China\\
$^5$Faculty of Science and Technology,Federation University Australia,Melbourne Australia\\
Email: \{liuyuan,wangyuancheng,linan95,xcheng,yfz,huangym\}@seu.edu.cn, guojun.lu@federation.edu.au}
}

\maketitle
\begin{abstract}
The main challenge of single image super resolution (SISR) is the recovery of high frequency details such as tiny textures. 
However, most of the state-of-the-art methods lack specific modules to identify high frequency areas, causing the output image to be blurred.
We propose an attention-based approach to give a discrimination between texture areas and smooth areas.
After the positions of high frequency details are located, high frequency compensation is carried out.
 This approach can incorporate with previously proposed SISR networks. By providing high frequency enhancement, better performance and visual effect are achieved. We also propose our own SISR network composed of DenseRes blocks. The block provides an effective way to combine the low level features and high level features.
 Extensive benchmark evaluation shows that our proposed method achieves significant improvement over the state-of-the-art works in SISR.
 \begin{comment}
Our proposed method achieves significant improvements over the state-of-the-art works in terms of peak signal to noise ratio (PSNR) and the structural similarity (SSIM) index. In addition, visual improvements in our results are
easily noticeable.
As for PSNR, we get an improvement of 0.54 dB and 0.52dB respectively over VDSR and DRCN.
\end{comment}
\end{abstract}

\begin{comment}
\begin{IEEEkeywords}
attention mechanism, DenseRes block, super resolution
\end{IEEEkeywords}
\end{comment}

\section{Introduction}\label{sec:Introduction}
The task of single image super-resolution (SISR) is to infer a high resolution (HR) image from a single low resolution (LR) input image. It is a highly ill-posed problem because the high frequency information such as tiny textures is lost during low-pass filtering and down-sampling. Thus, SISR is a one-to-many mapping. Our task is to find the most plausible HR image which recovers the tiny textures as close as possible.

In order to recover HR images from LR images, large receptive field is needed to take more contextual information from LR images. Using deeper networks is a better way to increase the receptive field. 
One drawback of deep networks is the vanishing-gradient problem which makes the network difficult to train. He et al. \cite{resnet} use the residual learning framework to ease the training of networks. Skip connections are another solution to boost the flow of gradient and information through the network. The low level features contain effective information and can be used to reconstruct the HR image. SISR will benefit from the collective information at different levels. 

\begin{comment}
\begin{figure}[htbp]
\centering
\includegraphics[width=9 cm]{multiple.png}
\caption{$\times$3 Super resolution results of our method on the image from BSD100 \cite{bsd300}. When attention is wiped off and the interpolation is not added to the final result of the network, the test result and the corresponding residual image is shown in (a) and (b). When the interpolation is added, the corresponding images is shown in (c) and (d). (e) and (f) is the results of the proposed method including attention mechanism and interpolation. 
When the attention mechanism is added, the visual effect is improved and much high frequency details are recovered. The boxplots shows the distribution of residual images. The boxplots of (b), (d) and (f) is on the left, in the middle and on the right of figure (g) respectively, which also shows the excellence of the proposed method.}
\label{fig:architecure of network}
\end{figure}
\end{comment}

The difficulty of SISR is the recovery of high frequency details such as tiny textures.
Mean squared error (MSE) between the output image and the original image is often applied as a loss function to train the convolutional neural network. However, in the process of pursuing high peak signal-to-noise ratio (PSNR), MSE will return the mean of many possible solutions and thus the output image will look blurry and implausible. In order to recover the high frequency details, perceptual losses \cite{perceptual_losses_for} have been proposed which encourage the network to produce images whose feature representations are similar, resulting in sharper images. Ledig at al. \cite{enhancenet_single,photo_real_s} combine adversarial network, perceptual loss and texture loss to encourage the output image to recover high frequency details such as tiny textures. But all these networks don't explicitly know the positions of high frequency details and they just try to restore the textures blindly. Thus, the performance of these network is not satisfactory. 

To solve these problems, first, based on denseNet \cite{densenet} which connects  each layer to every subsequent layer,  we propose a novel block called DenseRes block composed of residual building blocks (Resblock) \cite{resnet}. The output of every Resblock is connected to  every other Resblock, boosting the flow of information and avoiding  the re-learning of redundant features. With the DenseRes block, the gradient vanishing problem is alleviated and the network is easy to train.
Second, we provide an attention mechanism to cope with the recovery of high frequency details.
Inspired by U-net \cite{unet} which is used for semantic pixel-wise segmentation, we propose a novel hybrid densely connected U-net to help the network to discriminate if the areas are full of tiny textures in need of repairment or similar to the interpolation image. It works as a feature selector which selectively enhance the high frequency features. Thus, the textures can be restored as close as possible.

It is the first time that the attention mechanism is introduced into SISR. The method is simple and effective. By selectively providing high frequency enhancement, it alleviates the problem that output images tend to be blurred.
The attention mechanism can incorporate with previously proposed SISR networks. Higher PSNR and SSIM are achieved. 
Another contribution is that we propose the DenseRes block which provides an efficient way to combine the low level features and high level features. It's beneficial for the recovery of high frequency details.
\begin{comment}
Overall, the contributions of our method are mainly in two aspects:
\begin{enumerate}[1)]
    \item We propose a novel block called DenseRes block which provide an efficient way to combine low level features and high level features.
    \item Attention mechanism is proposed in order to give a discrimination between texture areas and smooth areas. Then high frequency enhancement is carried out. Therefore, high frequency details are restored as close as possible.
\end{enumerate}
\end{comment}

We evaluate our model on four publicly available benchmark datasets. It outperforms the current state-of-the-art approaches in terms of PSNR and the structural similarity (SSIM) index. As for PSNR, we get an improvement of 0.54 dB and 0.52dB respectively over VDSR \cite{accurate_image_s} and DRCN \cite{DRCN}. 

The remainder of this paper is organized as follows: related work which includes the algorithms for super resolution (SR) and attention mechanism is presented in Section \rom{2}, followed by the proposed network structure in Section \rom{3}. The experimental results and visual comparison with state-of-the-art results are provided in Section \rom{4}. We make a conclusion in Section \rom{5}.

\section{Related work}
\subsection{SISR}
Early approaches such as bicubic and Lanczos \cite{lanczos_filter} are easy to implement and the speed is high. But these methods often produce blurry results, lacking high frequency details. Many powerful methods such as sparse coding \cite{an_algor_for}were proposed to establish a complex mapping between low resolution and high resolution images. Sparse coding \cite{image_super_sparse,dis_object_tra} is based on the assumption that the sparse representation of the LR image over the LR dictionary is the same as that of the corresponding high resolution image over the HR dictionary. 

Recently, algorithms based on convolutional neural networks (CNN) have got excellent results and outperform other algorithms. Dong et al. \cite{learning_a_deep} upscaled an input image with bicubic interpolation and then trained a shallow convolutional network end-to-end to learn a nonlinear mapping from the LR input to a super-resolution output. Subsequently, various works \cite{accurate_image_s,DRCN,enhancenet_single} have successfully used deep networks in SISR and get higher PSNR values compared with shallow convolutional architectures. Recently, Lim et al. \cite{edsr} get the best results in the NTIRE2017 Super-Resolution Challenge \cite{ntire}. The depth of their network is up to 32.

\begin{comment}
The dense convolutional network (DenseNet) \cite{densenet} have several compelling advantages including alleviating the problem of vanishing-gradient, strengthening the flow of information and enormously reducing the number of parameters through feature reuse. Methods \cite{memnet,dense_skip} based on DenseNet have got compelling performance. In our studies, we also use the dense block.
\end{comment}

In many deep learning algorithms for SISR, the LR image is upsampled via bicubic interpolation as the input of the network \cite{accurate_image_s, DRCN}. This means that the SISR operation is performed in high resolution space, which is sub-optimal and adds computational complexity. Instead of an interpolated image, sub-pixel convolution layers \cite{real-time} are applied to upsample the feature maps to the size of the ground truth in the later layers of the network. This can reduce computations while the model capacity is reserved.

\begin{comment}
PixelRNN and PixelCNN \cite{pixel_recu,con_image} are probabilistic generative models that the probability of subsequent pixels are conditioned on previously observed pixels. A pixel recursive super resolution model \cite{pixel_recursive} , which is an extension of PixelCNNs, was proposed to address the task of high magnification factor.
\end{comment}

\subsection{Attention}

Methods based on attention mechanism have shown good performance on a range of tasks. In the field of speech recognition, an attention-based recurrent network decoder is used to transcribe speech utterances to characters \cite{listen_attend}. Chorowski et al. \cite{attention_based} improve the robustness to long inputs of speech with the attention mechanism. Hou et al. \cite{gaussian_pre} propose a simple but effective attention mechanism to achieve online speech recognition. In the field of machine translation, Ashish et al. \cite{attention_is} propose a new simple network based solely on attention mechanisms, dispensing with recurrence and convolutions entirely, which shows superior quality in machine translation tasks. Other works \cite{neural_mac,temp_atten} also achieve decent performance with the assistance of attention. In the field of computer vision, the attention mechanism has been used in image generation \cite{draw,generate_img} and image caption generation \cite{show_attend}. Yao et al. \cite{des_videos} propose a temporal attention mechanism to automatically select the most relevant temporal segments in the task of video description. As for salient object detection which aims to identify and locate distinctive regions that attract human attention, Zhang et al. \cite{salient_loss} design a symmetrical fully convolutional network to extract saliency features. Li et al. \cite{salient_weak} use weakly supervised methods and achieve comparable results with strongly-supervised methods. 
\begin{comment}
As for image classification, 
SEnet \cite{senet}, which uses global average pooling to generate channel-wise statistics and then give the weight for every channel, won the first place in the ILSVRC 2017 classification task. 
\begin{comment}
However, as far as we know, we are the first to introduce the attention mechanism into SISR.

\begin{comment}
As for the task of SISR, semantic segmentation networks \cite{unet,segnet,deeplab} can be used as attention mechanism to give a discrimination between texture areas and smooth areas, which can boost the recovery of high frequency details.
\end{comment}

\begin{comment}
As for the task of SISR, semantic segmentation networks can be used as attention mechanism to give a discrimination between texture areas and smooth areas. U-net \cite{unet} combine low-level features and high-level features via skip connections to alleviate information loss. Different from U-net, SegNet \cite{segnet} upsamples its input feature maps using the memorized max-pooling indices. DeepLab \cite{deeplab} uses atrous convolution to enlarge the field-of-view to avoid using max-pooling and strided convolution.
\end{comment}

\begin{figure*}[htbp]
\centering
\includegraphics[width=17 cm]{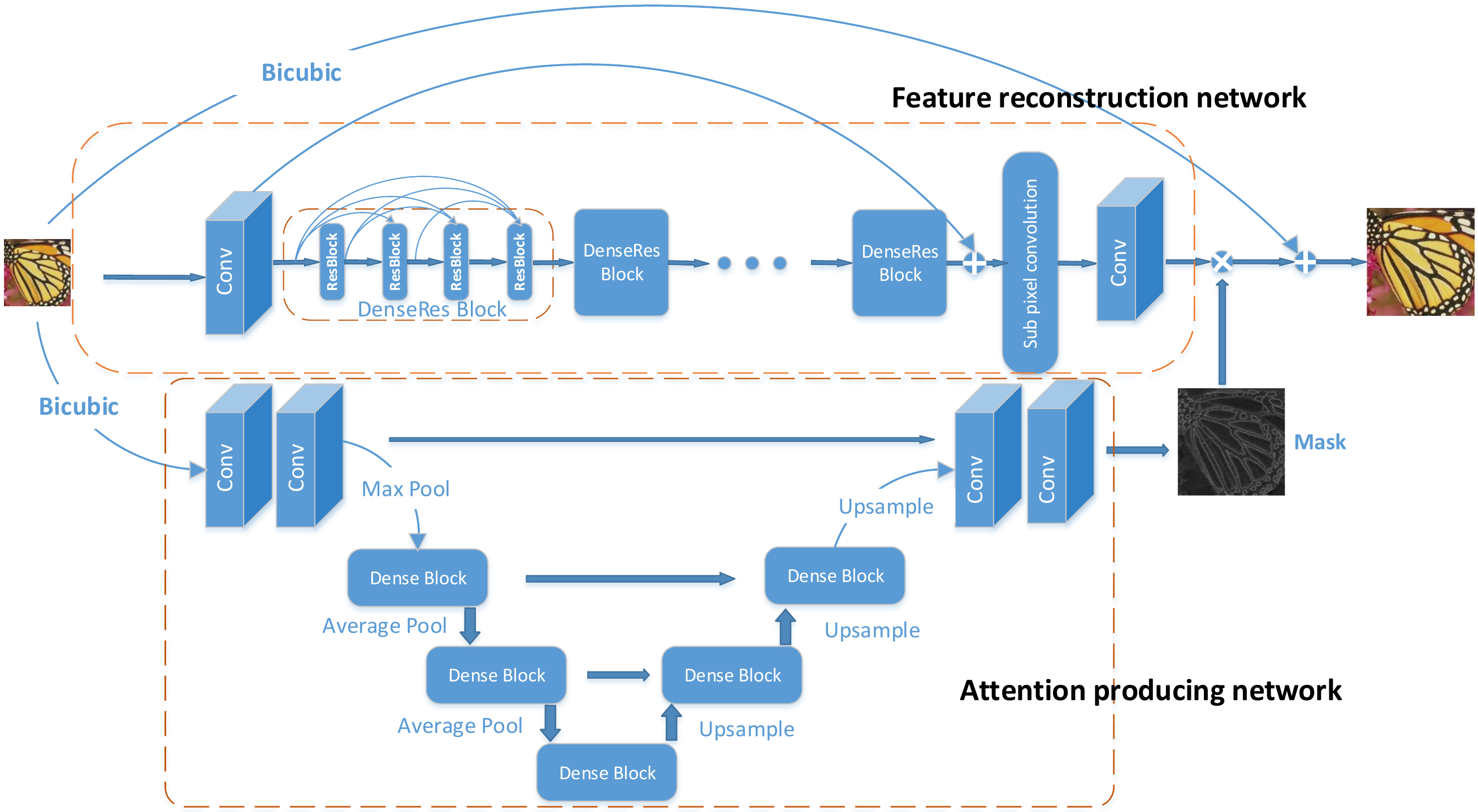}
\caption{The network architecture of our method. It consists of two parts: the feature reconstruction network which aims to recover the HR image and the attention producing network whose purpose is to selectively enhance high frequency features.}
\label{fig:architecure of network}
\end{figure*}

\section{Proposed Method}
In this section, we describe the proposed model architecture for SISR. The network aims to learn an end-to-end mapping function F between the LR image and the HR image. As shown in Fig. 1, our network is composed of two parts: the feature reconstruction network which aims to recover the HR image and the attention producing network whose purpose is to find high frequency details to be repaired. By the multiplication of the two networks' output, we will get the residual of the HR image.
 
\subsection{Feature reconstruction network}
 The feature reconstruction network aims to recover high-frequency details and reconstruct the high resolution image. It is a fully convolutional network. $L_{i-1}$ denotes the input of the $i^{th}$ convolutional layer, then the output of the $i^{th}$ layer is expressed as:
 \begin{equation}
L_{i}= \sigma (W_{i} \ast L_{i-1}+b_{i})
 \end{equation}
Where $\ast$ refers to the operation of convolution, $\sigma$ refers to the activation of rectified linear units (ReLU) \cite{relu}, $W_{i}$ and $b_{i}$ refer to the weights and biases of $i^{th}$ layer seperately.

The feature reconstruction network consists of three parts: a convolutional layer for feature extraction, multiple stacked DenseRes blocks  and a subpixel convolution \cite{real-time} layer as an upsampling module, which is illustrated in Fig. 1. Transition layers between DenseRes blocks are omitted for simplicity. In order to recover HR images from LR images, large receptive field is needed to take more contextual information to predict pixels in HR images. By the cascade of multiple DenseRes blocks, our network is deep and can make use of more pixels to get better performance in SISR.

DenseRes block is an important component of the feature reconstruction network. We now present the details of the block. The DenseRes block consists of residual building blocks(Resblock) \cite{resnet}, which show powerful learning ability for object recognition. Let $H_{i}$ be the input of the $i^{th}$ resblock, the output $F_{i}$ can be expressed as:
   \begin{equation}
 F_{i} = \Phi_{i}(H_{i},W_{i})+H_{i}
\end{equation}
where $W_{i}$ is the weight set to be learned in the $i^{th}$ Resblock and $\Phi_{i}$ is the Resblock function. Resblock includes two convolutional layers. Specifically, the Resblock function can be expressed as follows:
\begin{equation}
\Phi_{i}(H_{i},W_{i}) =\sigma_{2}(W^{2}_{i}\ast\sigma_{1}(W^{1}_{i}\ast H_{i}))
\end{equation}
where $W^{1}_{i}$ and $W^{2}_{i}$ are the weight of the two convolutional layers respectively and the bias is omitted for simplicity. $\sigma_{1}$ represents the batch normalization \cite{batch_norm} followed by the activation of ReLU \cite{relu}. $\sigma_{2}$ represents the batch normalization.

The DenseRes block includes several Resblocks. The input of the i$^{th}$ Resblock H$_{i}$ is the concatenation of the previous Resblocks' output. In typical feedforward CNN process, the high frequency information is easy to get lost at latter layers, and dense connections from previous Resblocks can alleviate such loss and further enhance high frequency signals.
However, if a large number of feature maps are directly fed into the subsequent Resblock, the model size and the computional size will be intensively increased. Thus, we use a convolutional layer with 1$\times$1 kernel to control how much of the previous states should be reserved. It adaptively learns the weights for different states. The input of the $i^{th} $ Resblock is expressed as:
   \begin{equation}
H_{i} = \sigma_{0}(W_{i}^{0}\ast[F_{1},F_{2},...,F_{i-1}])
\end{equation}
where $W_{i}^{0}$ is the weight of the convolutional layer with 1$\times$1 kernel and $\sigma_{0}$ represents the batch normalization followed by the activation of ReLU.
\subsection{Attention producing network}
The difficulty of SISR is the recovery of tiny textures. In the work of \cite{deep_edge}, they use the sobel operator to extract high-frequency components such as edges, then combine the high-frequency details with the LR image as input of the network, which is helpful for the recovery of shaper edges. But one disadvantage of this method is the hand-crafted features are not robust and thus the performance is not satisfactory.

If the network knows the exact locations of the tiny textures, it can give enhancement to features of these areas. Thus, more high frequency details will be recovered. The attention producing network can implement such functionality by providing attention mechanism which needs a large receptive field. With the architecture inspired by Unet \cite{unet} used for semantic segmentation, which is an encoder-decoder structure, the attention producing network can make use of a large region to provide attention.
 As shown in Fig. 1, the network consists of a contracting path(left side), an expansive path(right side) and skip connections. 
It takes an interpolated LR image (to the desired size) as input. The increased redundancy from interpolation can reduce the information loss in the forward propagation and thus is beneficial for a precise segmentation between texture areas and smooth areas. On the contrary, if we just take the LR image as input, the attention mechanism cannot perform well.

When compared with U-net, we substitute the convolution layer for a dense block. In the structure of DenseNet, each layer is connected to all the subsequent layers. This strengthens the reuse of information and thus solves the problem of vanishing-gradient. In addition, By the reuse of features, DenseNet structure can substantially reduce the number of parameters. Thus, it is easy to train and requires less computation complexity and memory cost. 

In the contracting path, the interpolated image will be extracted low level features by convolutional layers firstly. Then max pooling is followed to reduce the dimension of data and get larger receptive field. We use pooling two times in the contracting path. In this way, the network can make use of a larger region to predict whether a pixel belongs to the high-frequency region or not. In the expansive path, deconvolution layer is included to upsample the previous feature maps. Low-level features contain much useful information and much is lost during the forward propagation. By combining the low-level features and high-level features in the expansive path, the output can give a precise segmentation of whether this area is the field with textures or not and need to be repaired by the feature reconstruction network. The feature channels of the network's output is 1 and the size is the same as the HR image. In the final layer, we use the activation of sigmoid to control the output ranging from 0 to 1. We call the output mask. If the probability that the pixels belong to the texture areas is higher, the mask values will be closer to 1 which means these pixels need given more attention. If not, the mask values will be closer to 0.

 \begin{figure*}[htbp]
 \centering
  \subfloat[HR]{
\label{6bicubic.png}
\begin{minipage}[t]{0.17\textwidth}
\centering
\includegraphics[width=1\textwidth]{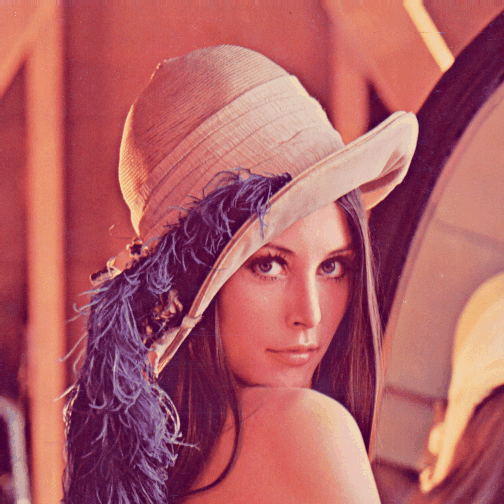}
\end{minipage}
}
  \subfloat[1 epoch]{
\label{6bicubic.png}
\begin{minipage}[t]{0.17\textwidth}
\centering
\includegraphics[width=1\textwidth]{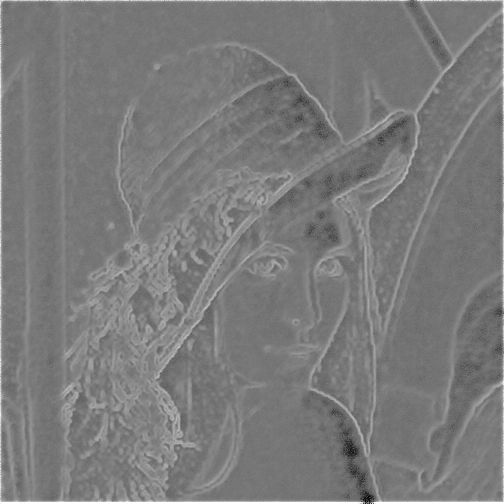}
\end{minipage}
}
  \subfloat[5 epochs]{
\label{6bicubic.png}
\begin{minipage}[t]{0.17\textwidth}
\centering
\includegraphics[width=1\textwidth]{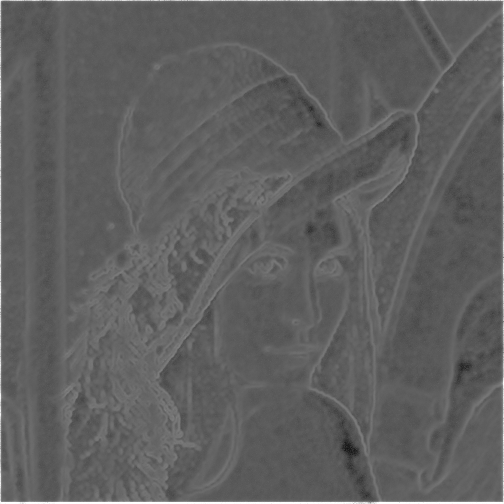}
\end{minipage}
}
  \subfloat[10 epochs]{
\label{6bicubic.png}
\begin{minipage}[t]{0.17\textwidth}
\centering
\includegraphics[width=1\textwidth]{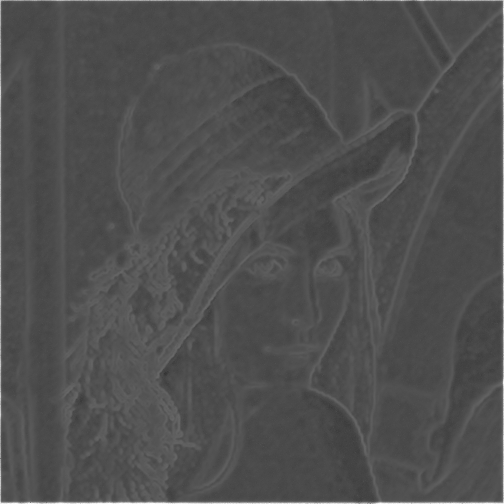}
\end{minipage}
}
  \subfloat[50 epochs]{
\label{6bicubic.png}
\begin{minipage}[t]{0.17\textwidth}
\centering
\includegraphics[width=1\textwidth]{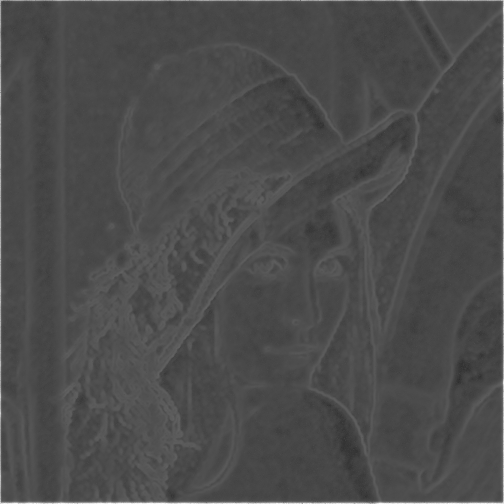}
\end{minipage}
}
\caption{Mask evolution during training. The upscaling factor is 3}
\end{figure*}

 \begin{figure*}[htbp]
 \centering
  \subfloat[Original image]{
\label{bsd3_gt_img.png}
\begin{minipage}[t]{0.17\textwidth}
\centering
\includegraphics[width=1\textwidth]{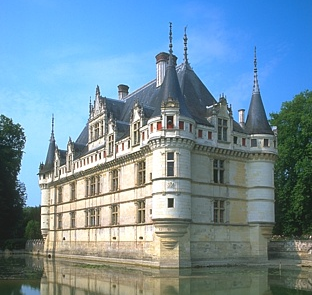}
\end{minipage}
}
  \subfloat[Mask]{
\label{bsd3_mask.png}
\begin{minipage}[t]{0.17\textwidth}
\centering
\includegraphics[width=1\textwidth]{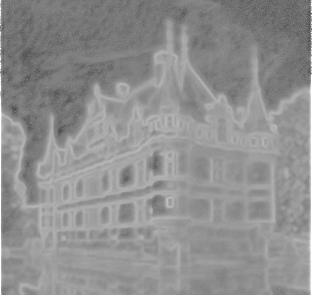}
\end{minipage}
}
  \subfloat[Residual before mask]{
\label{bsd3_origin.png}
\begin{minipage}[t]{0.17\textwidth}
\centering
\includegraphics[width=1\textwidth]{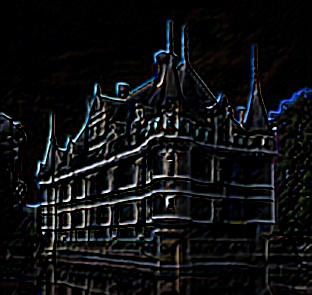}
\end{minipage}
}
  \subfloat[Residual after mask]{
\label{bsd3_with_mask.png}
\begin{minipage}[t]{0.17\textwidth}
\centering
\includegraphics[width=1\textwidth]{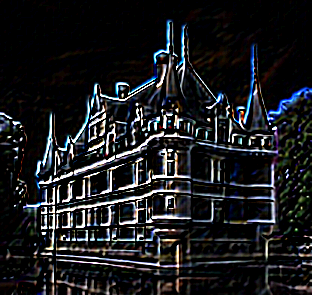}
\end{minipage}
}
  \subfloat[Ground truth]{
\label{bsd3_gt.png}
\begin{minipage}[t]{0.17\textwidth}
\centering
\includegraphics[width=1\textwidth]{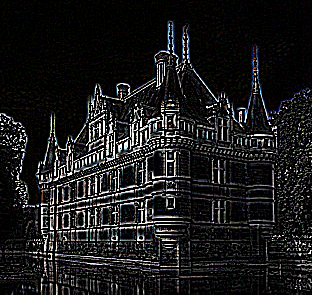}
\end{minipage}
}
\caption{Effect of the attention mechanism on the residual image. The upscaling factor is 3.}
\end{figure*}

\subsection{Residual learning of attention}
We get the residual of HR image by dot production of the output of the feature reconstruction network and the mask values. By adding the interpolated LR image, which is also the input of the attention producing network, the final HR result is achieved. It can be expressed as:
  \begin{equation}
HR^{c}(i,j) = F^{c}(i,j)\times M(i,j)+ILR^{c}(i,j)
\end{equation}
Where $F=[F^{1},F^{2},F^{3}]$ is the output of the feature reconstruction network, the number of output channels is 3. $M$ is the mask values. 
$ILR=[ILR^{1},ILR^{2},ILR^{3}]$ is the interpolated LR image.
$HR=[HR^{1},HR^{2},HR^{3}]$ is the final high resolution result of our method. $i$ and $j$ represent the pixel position in each channel and $c$ is the channel index. 
Thus, the attention producing network will encourage residual values from texture areas to be large, those not from texture areas to be close to 0.
The mask values $M$ works as a feature selector which enhance high frequency features and suppress noise.
 In the output images, the high frequency details will be recovered, and in those smooth places, noise will be removed. 
 
 \begin{comment}
 In forward inference, the attention mechanism serves as a feature selector. In addition, it also serves as a gradient update filter during back propagation. When updating parameters of the feature reconstruction network, the gradient can be expressed as :
 \begin{equation}
\(\frac{\partial M_{\phi}F_{\theta}}{\partial \theta}\) = M_{\phi}\(\frac{\partial F_{\theta}}{\partial \theta}\)
\end{equation}
where the $\theta$ are the parameters of the feature reconstruction network and the $\phi$ are the parameters of the attention producing network. Attention mechanism can prevent wrong gradients to update the parameters of the feature reconstruction network. Thus, our model is robust to noisy labels.
\end{comment}

\section{Experiments and Analysis}

\begin{table*}[htbp]
\tiny
  \caption{\upshape{Comparison of different methods including SelfEx  \cite{selfex}, SRCNN \cite{learning_a_deep}, A+ \cite{A+}, VDSR \cite{accurate_image_s} and DRCN \cite{DRCN}. Our models without the attention producing network is also included. The upscaling factor for super-resolution ranges from 2 to 4. The best performance is in bold.}}
\centering
 \label{tab:loop bound analysis}
 \begin{tabular}{c|c|c|c|c|c|c|c|c|c}
  \toprule
  Dataset& Scale &\tabincell{c}{Bicubic\\(PSNR/SSIM)}&\tabincell{c}{SelfEx\\(PSNR/SSIM)}&\tabincell{c}{SRCNN\\(PSNR/SSIM)}&\tabincell{c}{A+\\(PSNR/SSIM)} &\tabincell{c}{VDSR\\(PSNR/SSIM)}& \tabincell{c}{DRCN\\(PSNR/SSIM)} & \tabincell{c}{Proposed(no mask)\\(PSNR/SSIM)}&\tabincell{c}{Proposed\\(PSNR/SSIM)}\\
\midrule
  $Set5        $ & \tabincell{c}{$\times 2$ \\$\times 3$ \\$\times 4$} &\tabincell{c}{$33.66/0.9299$ \\$30.39/0.8682$ \\$28.42/0.8104$}& \tabincell{c}{$36.49/0.9537$ \\$32.58/0.9093$ \\$30.31/0.8619$}&  \tabincell{c}{$36.66/0.9542$ \\$32.75/0.9090$ \\$30.48/0.8628$} & \tabincell{c}{$36.54/0.9544$ \\$32.58/0.9088$ \\$30.28/0.8603$} &   \tabincell{c}{$37.53/0.9587$ \\$33.66/0.9213$ \\$31.35/0.8838$}&  \tabincell{c}{$37.63/0.9588$ \\$33.82/0.9226$ \\$31.53/0.8854$}& $\tabincell{c}{$37.76/0.9603$ \\$33.91/0.9220$ \\$31.65/0.8891$}$ &\tabincell{c}{$\bm{37.79/0.9608}$ \\$\bm{34.01/0.9231}$ \\$\bm{31.77/0.8972}$}\\
\midrule
  $Set14      $  &  \tabincell{c}{$\times 2$ \\ $\times 3$ \\$\times 4$}&$  \tabincell{c}{$30.24/0.8688$ \\$27.55/0.7742$ \\$26.00/0.7027$}$ &   \tabincell{c}{$32.26/0.9040$ \\$29.05/0.8164$ \\$27.24/0.7451$}&\tabincell{c}{$32.45/0.9067$ \\$29.30/0.8215$ \\$27.50/0.7513$} & \tabincell{c}{$32.28/0.9056$ \\$29.13/0.8188$ \\$27.32/0.7491$}&  \tabincell{c}{$33.03/0.9124$ \\$29.77/0.8314$ \\$28.01/0.7674$}& \tabincell{c}{$33.04/0.9118$ \\$29.76/0.8311$ \\$28.02/0.7670$} & $\tabincell{c}{$33.16/0.9153$ \\$30.19/0.8625$ \\$28.21/0.7838$}$&\tabincell{c}{$\bm{33.45/0.9193}$ \\$\bm{31.21/0.8873}$ \\$\bm{28.43/0.7954}$}\\
\midrule
  $BSD100  $ & \tabincell{c}{$\times 2$ \\ $\times 3$ \\$\times 4$}&  $  \tabincell{c}{$29.56/0.8431$ \\$27.21/0.7385$ \\$25.96/0.6675$}$ & \tabincell{c}{$31.21/0.8863$ \\$28.29/0.7835$ \\$26.82/0.7087$}& \tabincell{c}{$31.36/0.8879$ \\$28.41/0.7863$ \\$26.90/0.7101$} & \tabincell{c}{$31.21/0.8863$ \\$28.29/0.7835$ \\$26.82/0.7087$} &  \tabincell{c}{$31.90/0.8960$ \\$28.82/0.7976$ \\$27.29/0.7251$}& \tabincell{c}{$31.85/0.8942$ \\$28.80/0.7963$ \\$27.23/0.7233$}& \tabincell{c}{$32.16/0.9007$ \\$29.01/\bm{0.8107}$ \\$27.26/0.7118$}&\tabincell{c}{$\bm{32.25/0.9021}$ \\$\bm{29.13}/0.8103$ \\$\bm{27.45/0.7248}$}\\
 \midrule
 $Urban100  $ & \tabincell{c}{$\times 2$ \\ $\times 3$ \\$\times 4$}&  $  \tabincell{c}{$26.88/0.8403$ \\$24.46/0.7349$ \\$23.14/0.6577$}$ &  \tabincell{c}{$29.20/0.8938$ \\$26.03/0.7973$ \\$24.32/0.7183$}& \tabincell{c}{$29.50/0.8946$ \\$26.24/0.7989$ \\$24.52/0.7221$} & \tabincell{c}{$29.20/0.8938$ \\$26.03/0.7973$ \\$24.32/0.7183$} &  \tabincell{c}{$30.76/0.9140$ \\$27.14/0.8279$ \\$25.18/0.7524$}& \tabincell{c}{$30.75/0.9133$ \\$27.15/0.8276$ \\$25.14/0.7510$} &\tabincell{c}{$31.61/0.8838$ \\$27.39/\bm{0.8401}$ \\$25.29/0.7613$}&\tabincell{c}{$\bm{32.36/0.9203}$ \\$\bm{27.53}/0.8339$ \\$\bm{25.55/0.7729}$}\\

  \bottomrule
 \end{tabular}
\end{table*}

\subsection{Datasets}
 As for training, we use images from the DIV2K dataset \cite{ntire}. It consists of 1000 diverse 2K resolution RGB images.There are 800 images for training, 100 images for validation and 100 images for test among them. During the evaluation, we perform experiments on four publicly available benchmark datasets including Set5 \cite{set5}, Set14 \cite{set14}, BSD100 and Urban100 \cite{selfex}. For the datasets of Set5 and Set14, they contain 5 and 14 images respectively. BSD100 is the testing set of the Berkeley segmentation dataset BSD300 \cite{bsd300}. Urban100 contains 100 images with a variety of real-world structures. Our experiments are performed with different scale factors varying from 2 to 4 between the low resolution and high resolution images. As for evaluation, we use PNSR and SSIM as metrics.
 
\subsection{Training details and parameters}
We use the RGB image patches whose sizes are 48$\times$48 as input and the ground truth are the corresponding HR patches whose sizes are 48r$\times$48r. r is the upscaling factor. We augment the training data with random horizontal flips and 90 rotations. All the images are from the DIV2K dataset. Before being fed into the network, the image values are normalized to the range between 0 and 1. As for the network, the filter size is set to 3$\times$3 in all weight layers. We use the method of Xavier proposed in \cite{xavier} to initialize the weights and the bias are initialized to zero. We choose the rectified linear units (ReLU) as the activation function. The attention producing network includes a contracting path and an expansive path. In the contracting path, a 2$\times$2  pooling operation with stride 2 is used for downsampling. In the expansive path, deconvolution layer is used to upsample the feature map. As for the feature reconstruction block, one DenseRes block includes 4 Resblocks and we use 6 DenseRes blocks in all.

The network is optimized end to end using Adam \cite{adam}. The batch size is 16 and the training will stop after 80 epochs when no improvement is observed. Initial learning rate is 0.0001 which will decrease by 50 percent for each ten epochs. Because the loss function of MSE will cause the output images to be blurry and lacking high-frequency details, L1 loss is used as the loss function which provides better convergence according to \cite{edsr}. Our implementation is based on Tensorflow \cite{tensorflow}.

\subsection{The importance of the attention producing network}
\begin{comment}
 The trained attention producing network is able to give a perfect segmentation of whether the pixel belongs to texture areas or not. We call its output mask. If the probability that the pixels belong to texture areas is high, the corresponding mask value will be close to 1. If not, the mask value will be close to 0. Because the HR image's residual originates from the multiplication of the output of the feature reconstruction network and the mask values, the attention producing network will encourage those residual values from texture areas to be large, those not from texture areas to be close to 0. Thus, in the output images, the high frequency details will be recovered, and in those places with little high frequency details, noise will be removed. 
 \end{comment}
 
 The feature reconstruction network and the attention producing network are trained jointly in the whole procedure. 
 We call the output of the attention producing network mask. From Fig. 2, which shows changes of the output masks in the training procedure, we can conclude that the mask will give a precise attention when the training is finished. Thus, the texture areas can be recovered and noise will be removed in the assistance of the mask. Fig. 3 illustrates that when mask is added, the residual image will be rich in texture information. 
 As illustrated in Fig. 5, with the help of attention producing network, tiny textures are recovered. In addition, the output image has better visual effect than the ground truth image due to high frequency enhancement of the attention mechanism.
 On the contrary, when the attention producing network is wiped off, the network cannot provide a better recovery of tiny textures. The quantitative results by PSNR and SSIM are presented in Table 1. Compared with our method without attention producing network, in terms of PSNR, our method with attention producing network gets an improvement of 0.368 dB on average. An increasement of 1.02 dB is even achieved for the dataset of Set14 on $\times$ 3 super resolution. As for SSIM, our method with the mask gets an improvement of 0.0111 over that without the mask.
Thus, we can come to the conclusion that the attention producing network is critical in the recovery of HR images. 

We also compare the attention mechanism with the work of \cite{deep_edge}, which use sobel operator to help the recovery of high frequency details. We substitute the attention producing network for the sobel operator while the feature reconstruction network remains unchanged. When the network finishes training, quantitative results show that a decline of 0.28 dB is achieved compared with the network with attention mechanism. Thus, our attention mechanism can give a better assistance of the feature reconstruction network.

  \begin{figure}[H]
\centering
\includegraphics[width=8 cm]{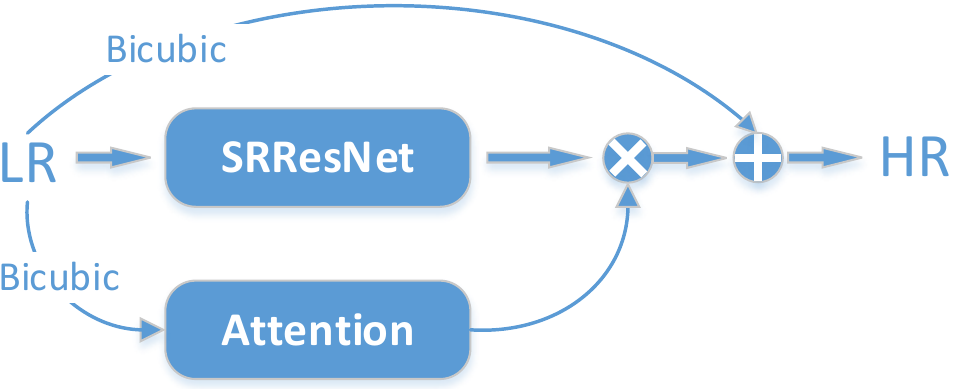}
\caption{The architecture of SRResNet with attention producing network.}
\label{fig:architecure of network}
\end{figure}

\begin{figure*}[htbp]
\centering
\includegraphics[width=18 cm]{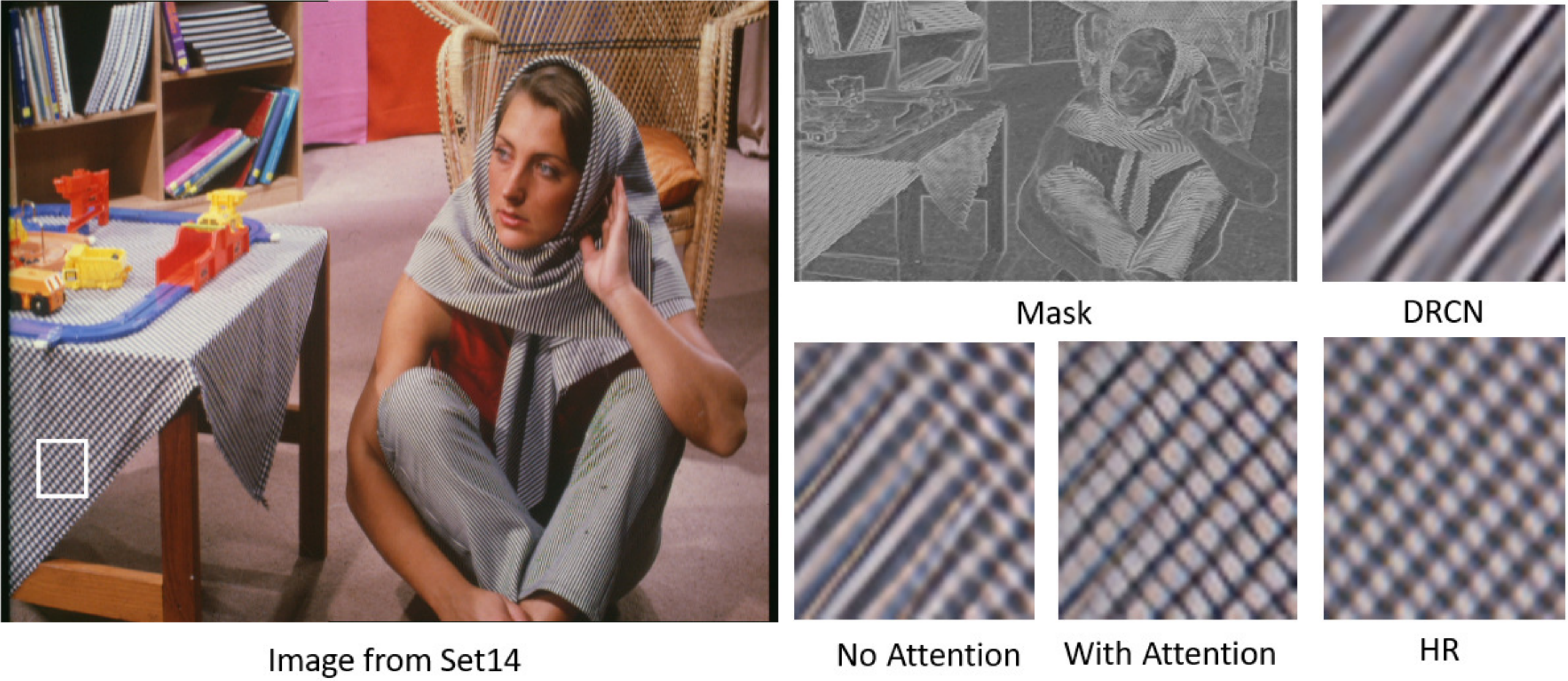}
\caption{ The comparison of our methods between with attention mechanism and without attention mechanism. The SR result with attention mechanims has even better visual effect than the ground truth due to high frequency enhancement.}
\label{fig:architecure of network}
\end{figure*}
%%%%%

 \begin{figure*}[htbp]
\centering

  \subfloat{
\label{18bicubic}
\begin{minipage}[t]{0.31\textwidth}
\centering
\includegraphics[width=1\textwidth]{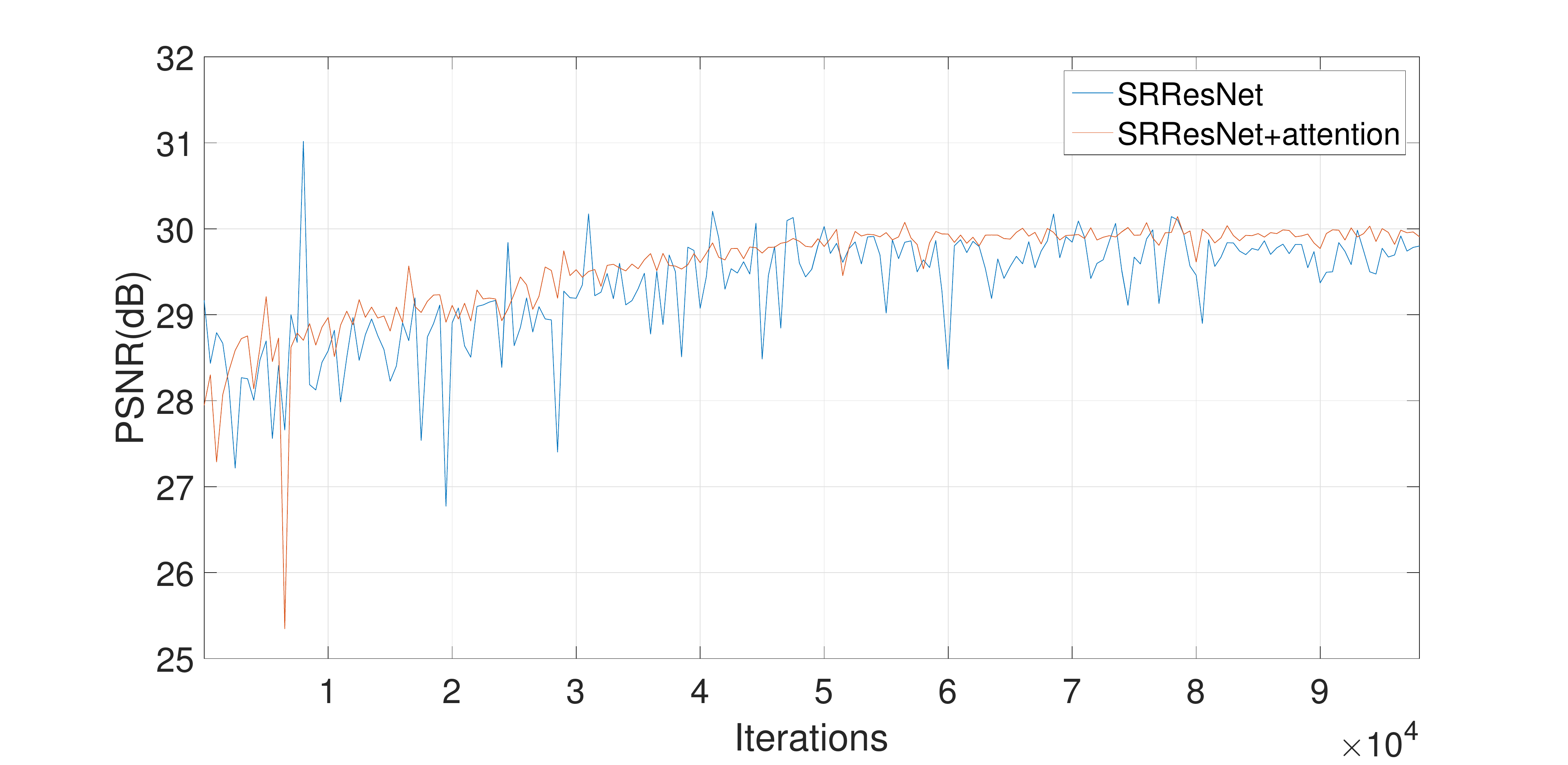}
\end{minipage}
}
  \subfloat{
\label{18bicubic}
\begin{minipage}[t]{0.31\textwidth}
\centering
\includegraphics[width=1\textwidth]{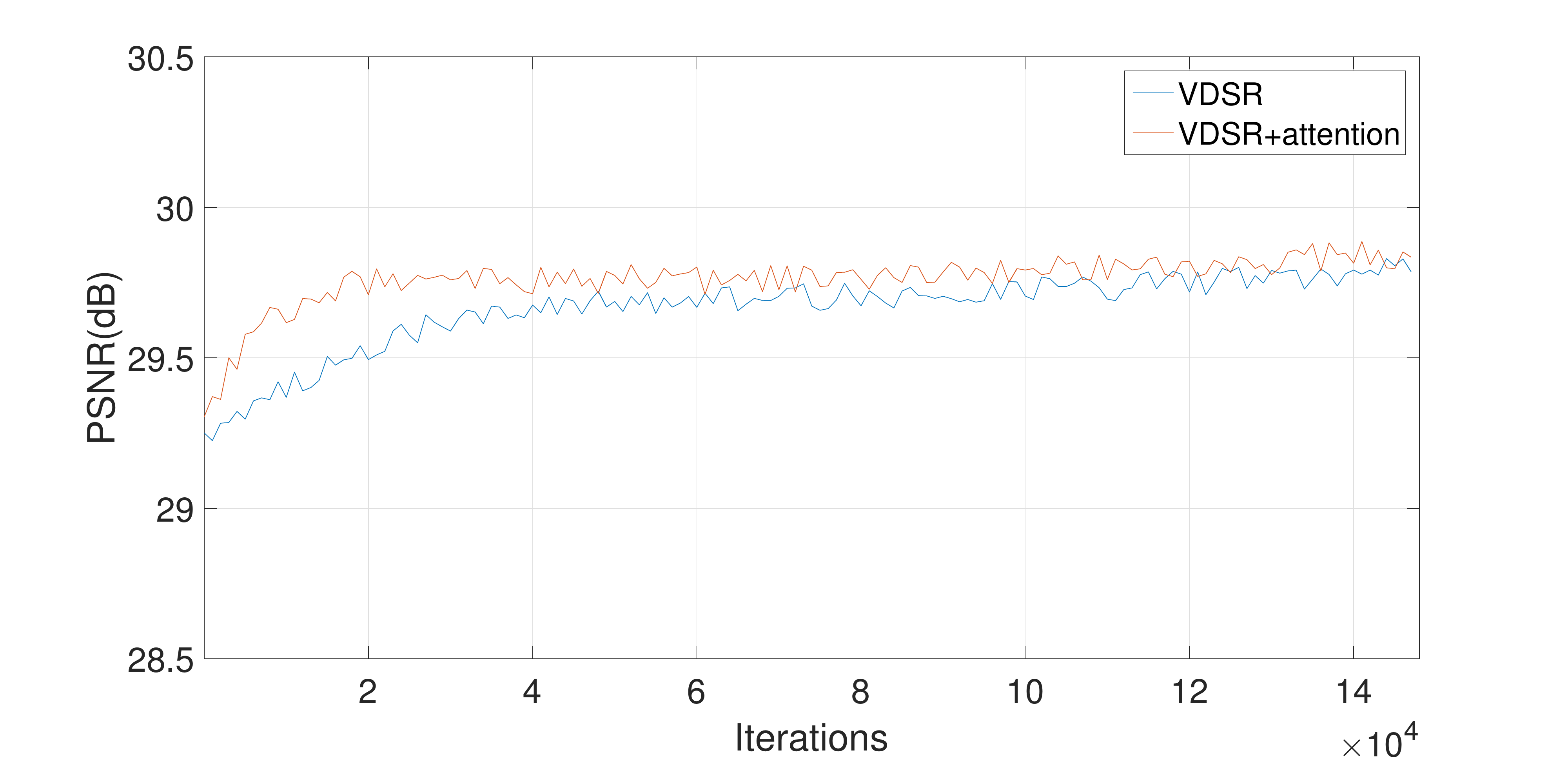}
\end{minipage}
}
  \subfloat{
\label{18bicubic}
\begin{minipage}[t]{0.31\textwidth}
\centering
\includegraphics[width=1\textwidth]{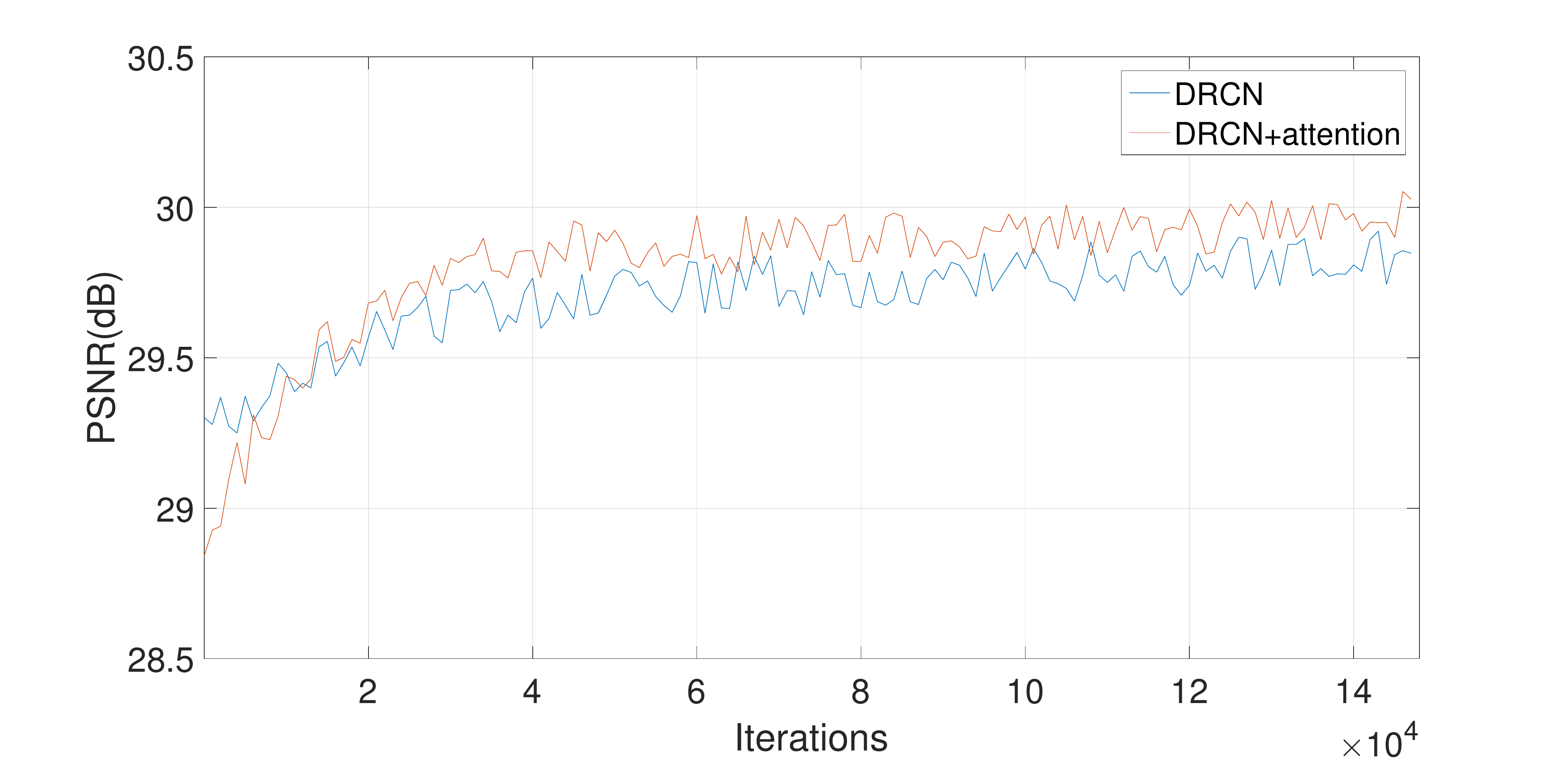}
\end{minipage}
}
\begin{comment}
 \subfig{
    \label{fig:1	8bicubic} %% label for first subfigure
    \includegraphics[width=5.5cm\textwidth]{SRResNet_mask_no.pdf}}
  \subfig{
    \label{fig:18srcnn} %% label for first subfigure
    \includegraphics[width=5.5cm\textwidth]{vdsr_mask_no.pdf}}
  \subfig{
    \label{fig:18drcn} %% label for first subfigure
    \includegraphics[width=5.5cm\textwidth]{DRCN_mask_no.pdf}}
    \end{comment}
\caption{Performance curve for SRResNet, VDSR and DRCN. Comparisons are made between the networks with attention mechanism and without attention mechanism. Two networks are tested under 'Set14' dataset with upscaling factor 3.}
\end{figure*}

\subsection{The flexibility of the attention producing network}
The attention producing network can be applied in other SISR networks. By providing attention, the attention producing network will improve the overall performance of other networks. To illustrate this point, we integrate  the attention mechanism into SRResNet \cite{photo_real_s} , which is shown in Fig. 4. In order to make a comparison, we train SRResNet without attention and SRResNet with attention from scratch. The two networks are trained with the same training sets. Training details are also the same. Performance curve 
is illustrated in Fig. 6. It's obvious that SRResNet with attention producing network has a better convergence and the test performance is 
better than SRResNet without attention producing network. When the two networks finish training, in terms of PSNR,  
SRResNet with attention is 0.15dB higher than SRResNet 
without attention on average, which is tested on the four publicly available benchmark datasets with upscaling factor 3. We also integrate the attention mechanism into VDSR \cite{accurate_image_s} and DRCN \cite{DRCN} . From Fig. 6, we can conclude that VDSR and DRCN have a better convergence when the attention mechanism is added. As for PSNR, we get increasement of 0.093 dB and 0.18dB respectively for VDSR and DRCN. Thus, the effectiveness of the attention mechanism is verified.

 \begin{figure*}[htbp]
 \centering
 \begin{comment}
    \subfiguret{
    \label{fig:18bicubic} %% label for first subfigure
    \includegraphics[width=0.15\textwidth]{small_18bicubic.pdf}}
  \subfigure{
    \label{fig:18srcnn} %% label for first subfigure
    \includegraphics[width=0.15\textwidth]{small_18srcnn.pdf}}
  \subfigure{
    \label{fig:18drcn} %% label for first subfigure
    \includegraphics[width=0.15\textwidth]{small_18drcn.pdf}}
  \subfigure{
    \label{fig:18vdsr} %% label for first subfigure
    \includegraphics[width=0.15\textwidth]{small_18vdsr.pdf}}
  \subfigure{
    \label{fig:18edsr_mask} %% label for first subfigure
    \includegraphics[width=0.15\textwidth]{small_18edsr_mask.pdf}}
  \subfigure{
    \label{fig:18hr} %% label for first subfigure
    \includegraphics[width=0.15\textwidth]{small_18hr.pdf}}
\end{comment}

  \subfloat{
\label{18bicubic}
\begin{minipage}[t]{0.15\textwidth}
\centering
\includegraphics[width=1\textwidth]{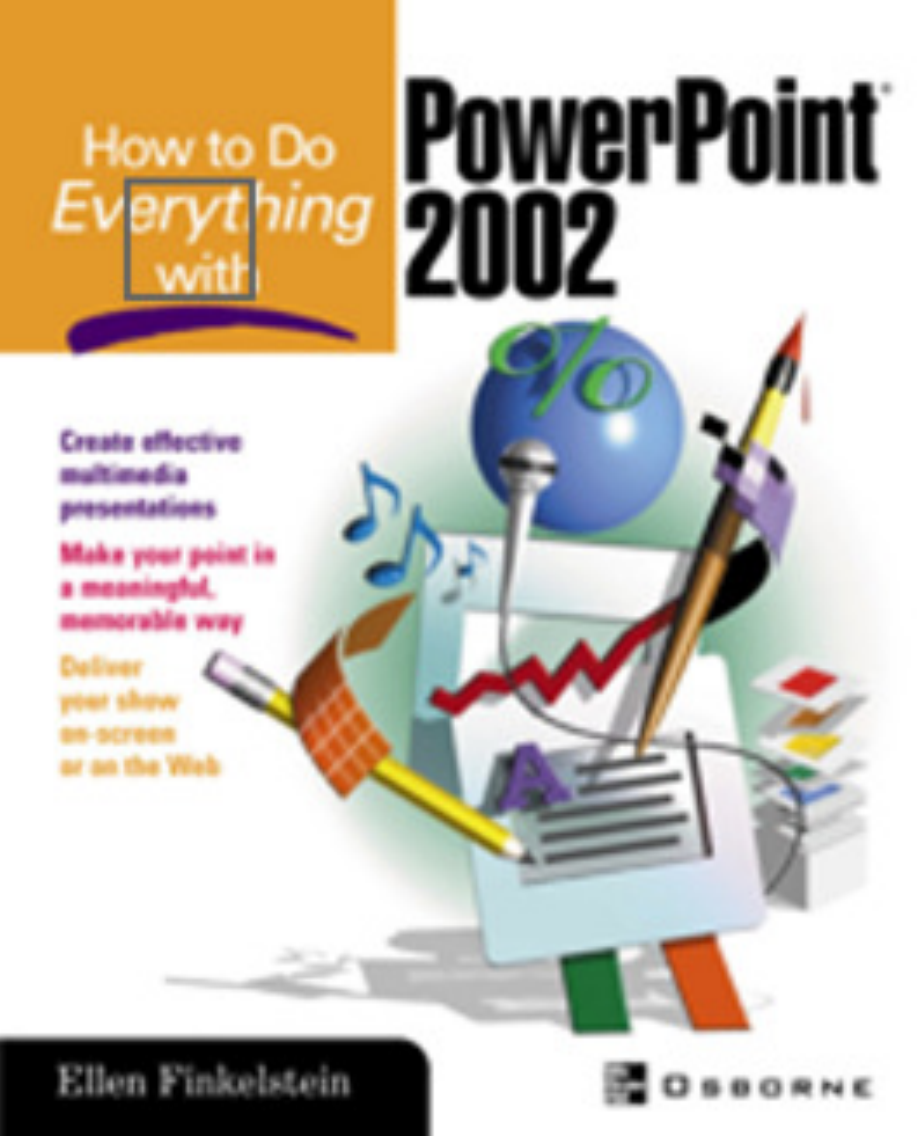}
\end{minipage}
}
  \subfloat{
\label{18bicubic}
\begin{minipage}[t]{0.15\textwidth}
\centering
\includegraphics[width=1\textwidth]{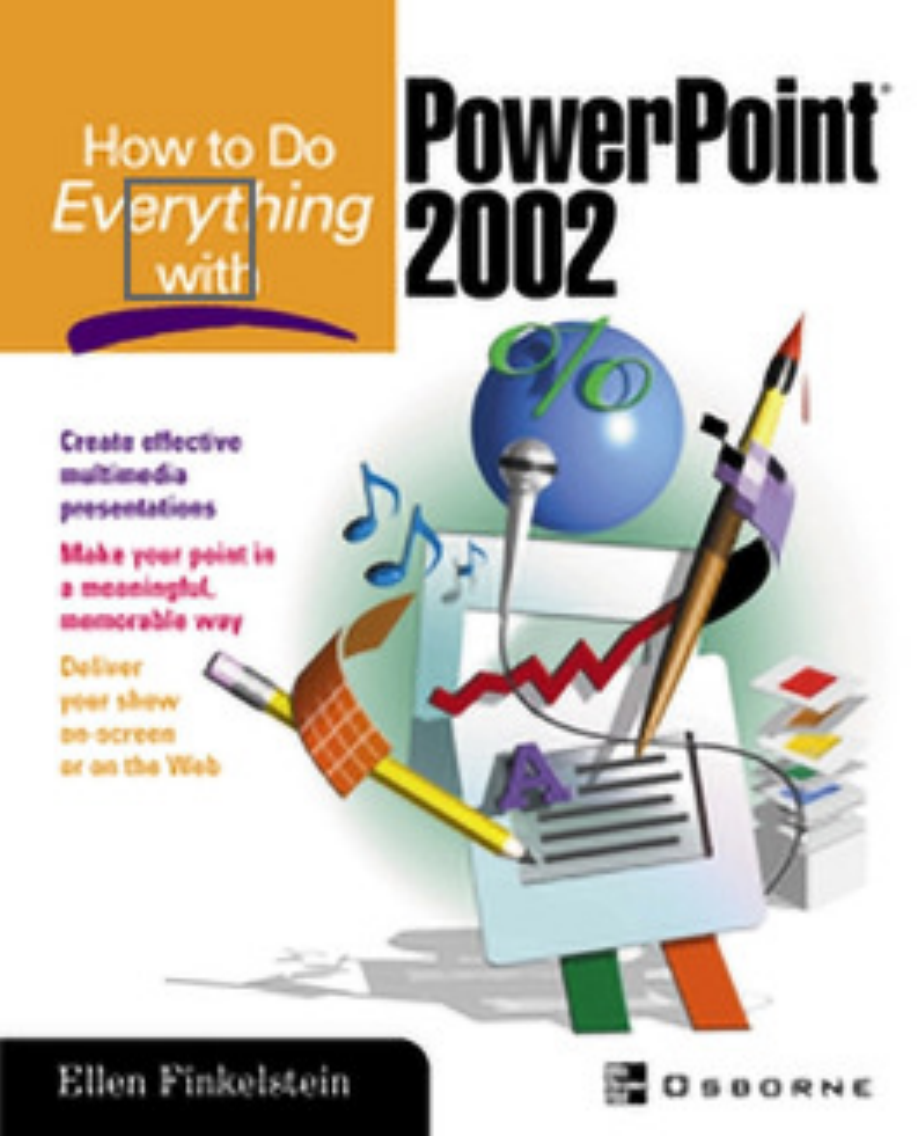}
\end{minipage}
}
  \subfloat{
\label{18bicubic}
\begin{minipage}[t]{0.15\textwidth}
\centering
\includegraphics[width=1\textwidth]{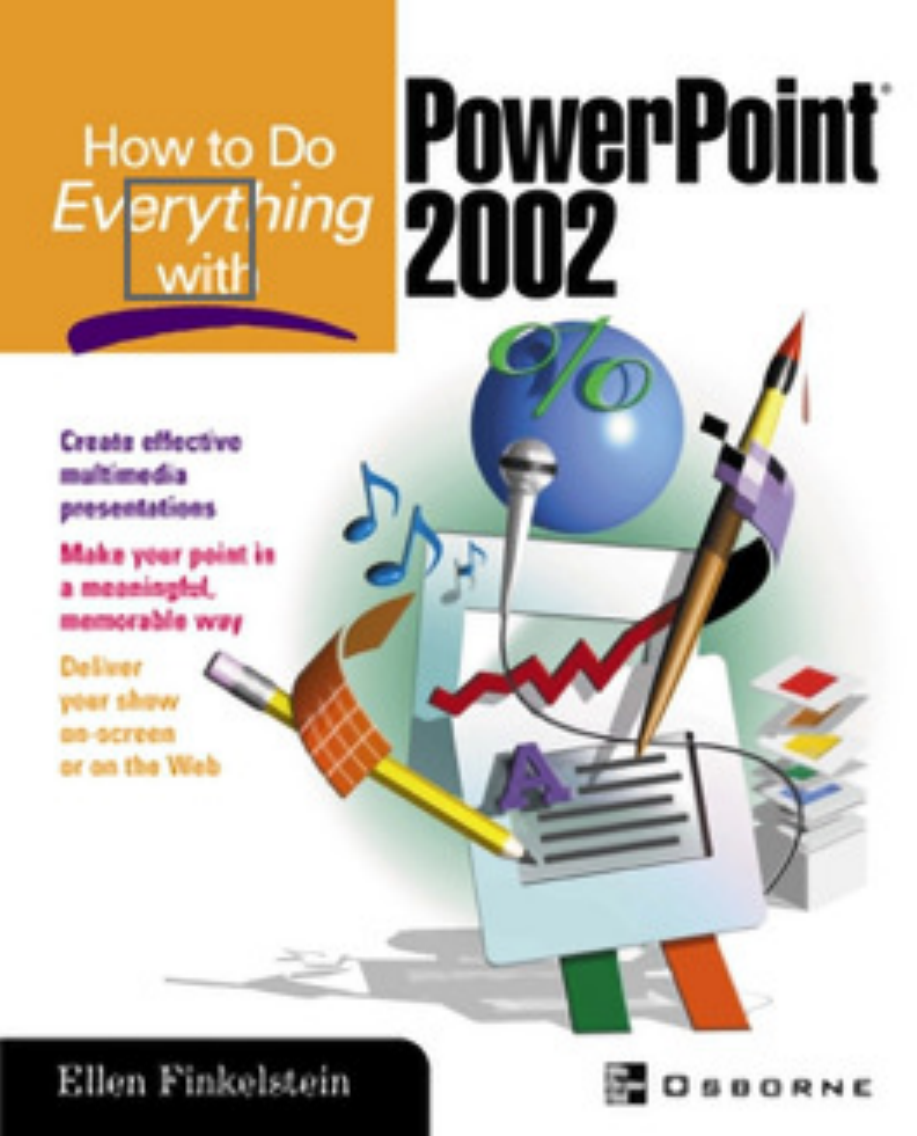}
\end{minipage}
}
  \subfloat{
\label{6bicubic.png}
\begin{minipage}[t]{0.15\textwidth}
\centering
\includegraphics[width=1\textwidth]{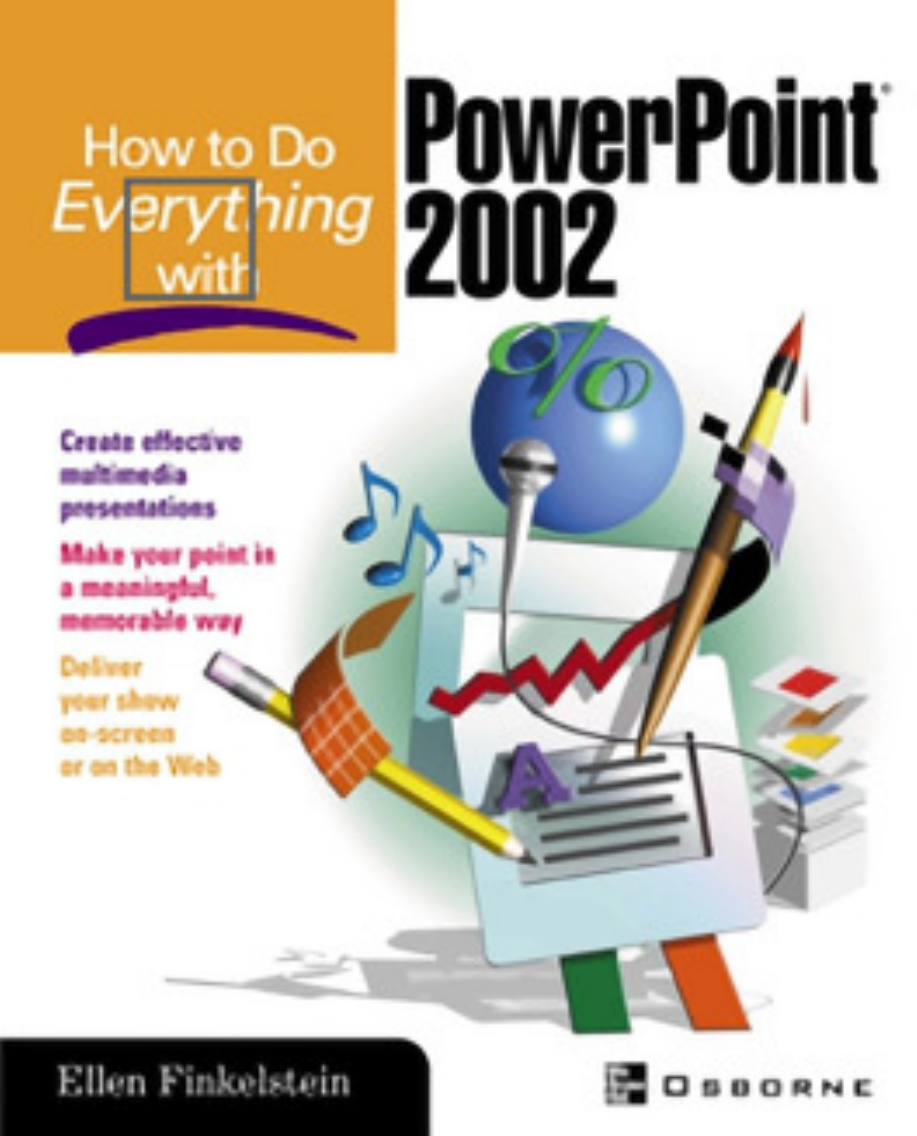}
\end{minipage}
}
  \subfloat{
\label{6bicubic.png}
\begin{minipage}[t]{0.15\textwidth}
\centering
\includegraphics[width=1\textwidth]{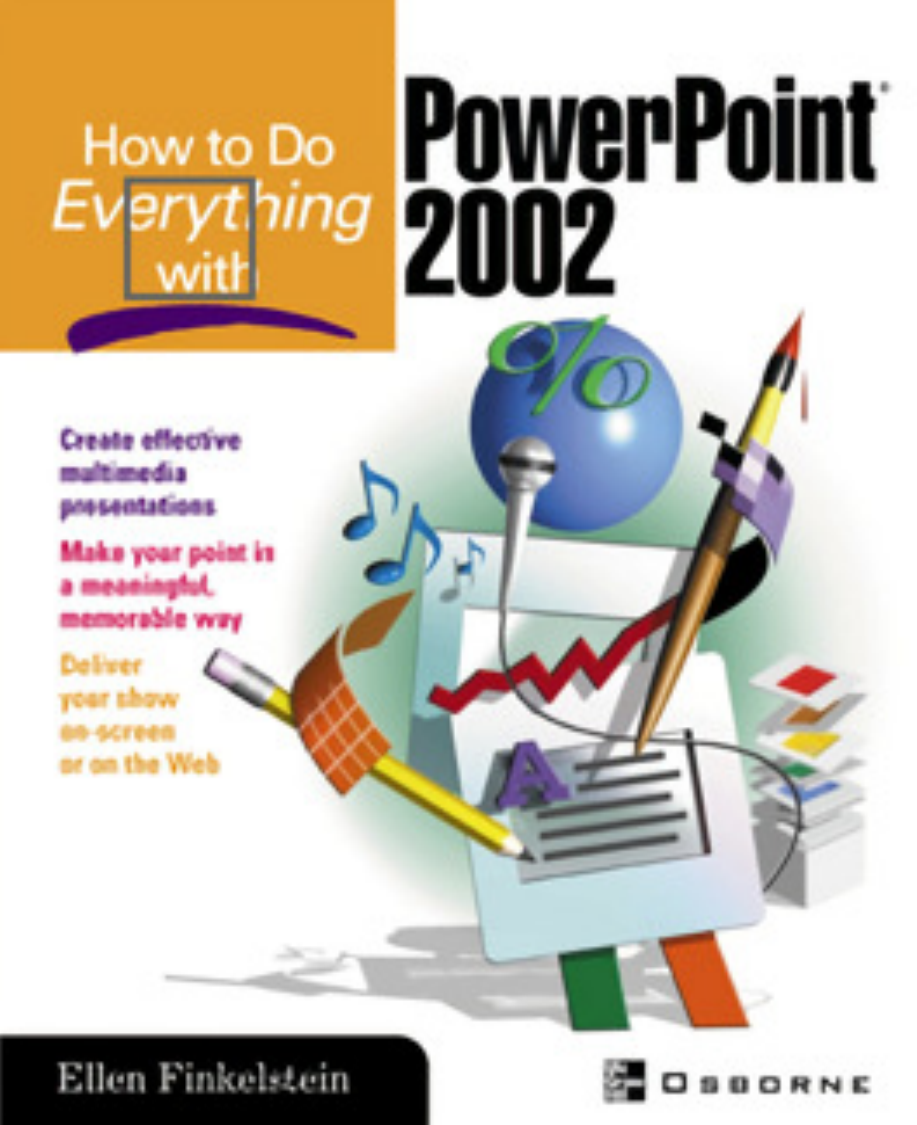}
\end{minipage}
}
  \subfloat{
\begin{minipage}[t]{0.15\textwidth}
\centering
\includegraphics[width=1\textwidth]{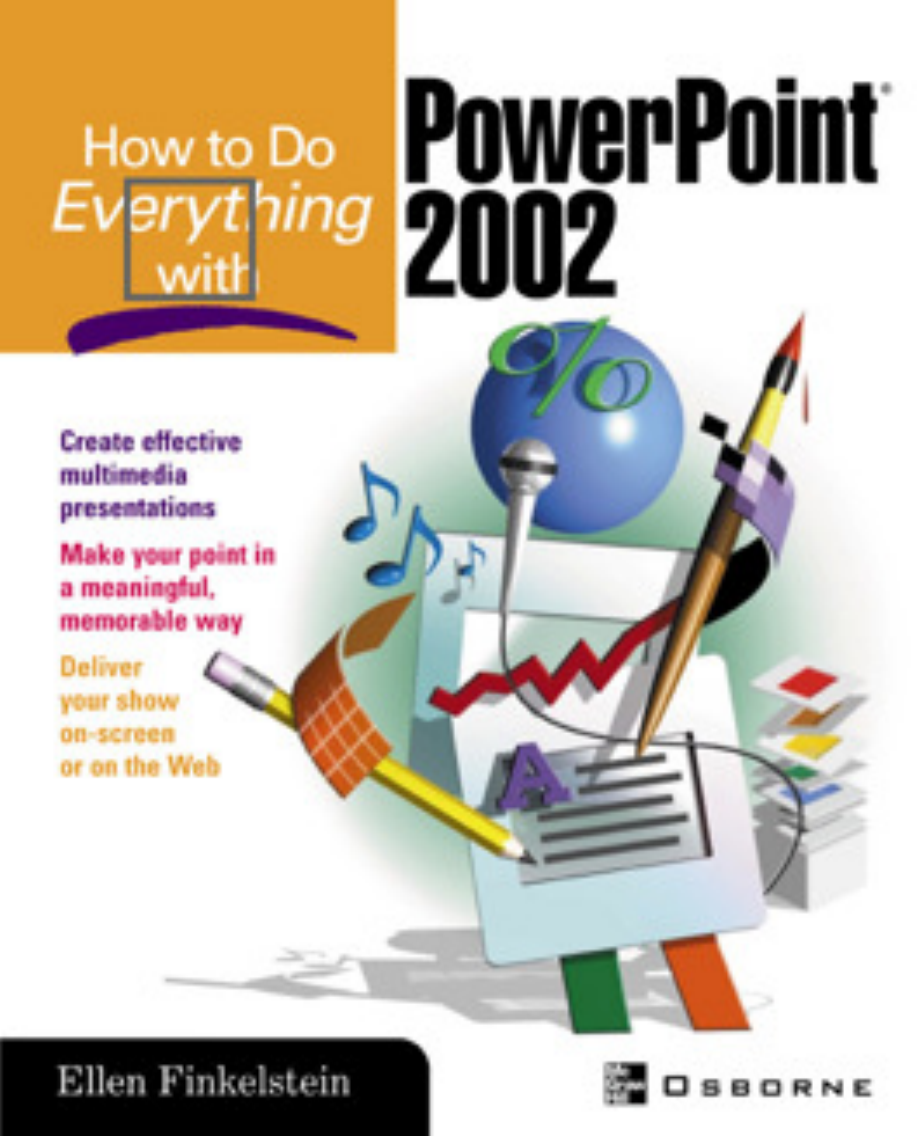}
\end{minipage}
}

  \subfloat[Bicubic]{
\label{6bicubic.png}
\begin{minipage}[t]{0.15\textwidth}
\centering
\includegraphics[width=1\textwidth]{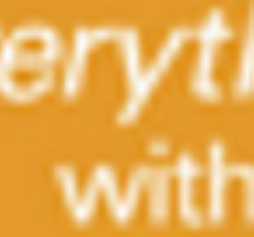}
\end{minipage}
}
  \subfloat[SRCNN]{
\label{6bicubic.png}
\begin{minipage}[t]{0.15\textwidth}
\centering
\includegraphics[width=1\textwidth]{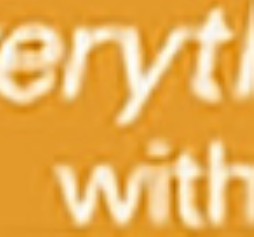}
\end{minipage}
}
  \subfloat[DRCN]{
\label{6bicubic.png}
\begin{minipage}[t]{0.15\textwidth}
\centering
\includegraphics[width=1\textwidth]{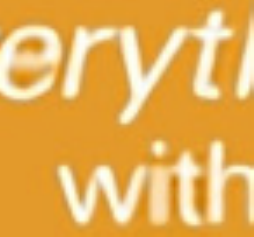}
\end{minipage}
}
  \subfloat[VDSR]{
\label{6bicubic.png}
\begin{minipage}[t]{0.15\textwidth}
\centering
\includegraphics[width=1\textwidth]{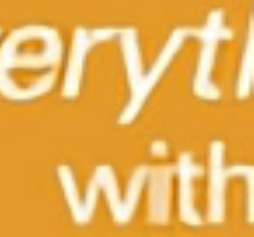}
\end{minipage}
}
  \subfloat[Proposed Method]{
\label{6bicubic.png}
\begin{minipage}[t]{0.15\textwidth}
\centering
\includegraphics[width=1\textwidth]{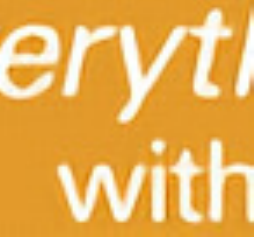}
\end{minipage}
}
  \subfloat[HR]{
\begin{minipage}[t]{0.15\textwidth}
\centering
\includegraphics[width=1\textwidth]{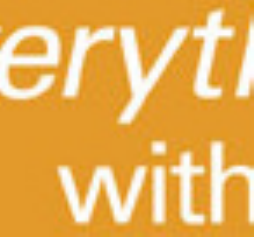}
\end{minipage}
} 
  
 \begin{comment}
      \subfigure{
    \label{fig:6bicubic} %% label for first subfigure
    \includegraphics[width=0.14\textwidth]{small_6bicubic.pdf}}
  \subfigure{
    \label{fig:6srcnn} %% label for first subfigure
    \includegraphics[width=0.14\textwidth]{small_6srcnn.pdf}}
  \subfigure{
    \label{fig:6drcn} %% label for first subfigure
    \includegraphics[width=0.14\textwidth]{small_6drcn.pdf}}
  \subfigure{
    \label{fig:6vdsr} %% label for first subfigure
    \includegraphics[width=0.14\textwidth]{small_6vdsr.pdf}}
  \subfigure{
    \label{fig:6edsr} %% label for first subfigure
    \includegraphics[width=0.14\textwidth]{small_6edsr_mask.pdf}}
  \subfigure{
    \label{fig:6hr} %% label for first subfigure
    \includegraphics[width=0.14\textwidth]{small_6hr.pdf}}
    \end{comment}
 %%%%%

   \subfloat{
\begin{minipage}[t]{0.15\textwidth}
\centering
\includegraphics[width=1\textwidth]{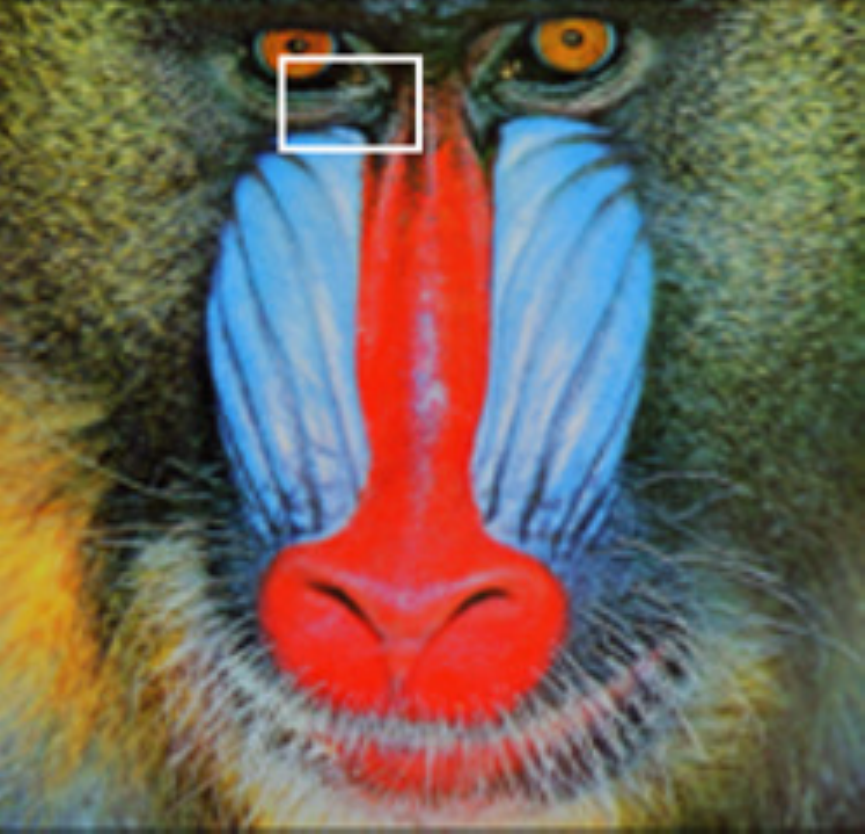}
\end{minipage}
}
  \subfloat{
\begin{minipage}[t]{0.15\textwidth}
\centering
\includegraphics[width=1\textwidth]{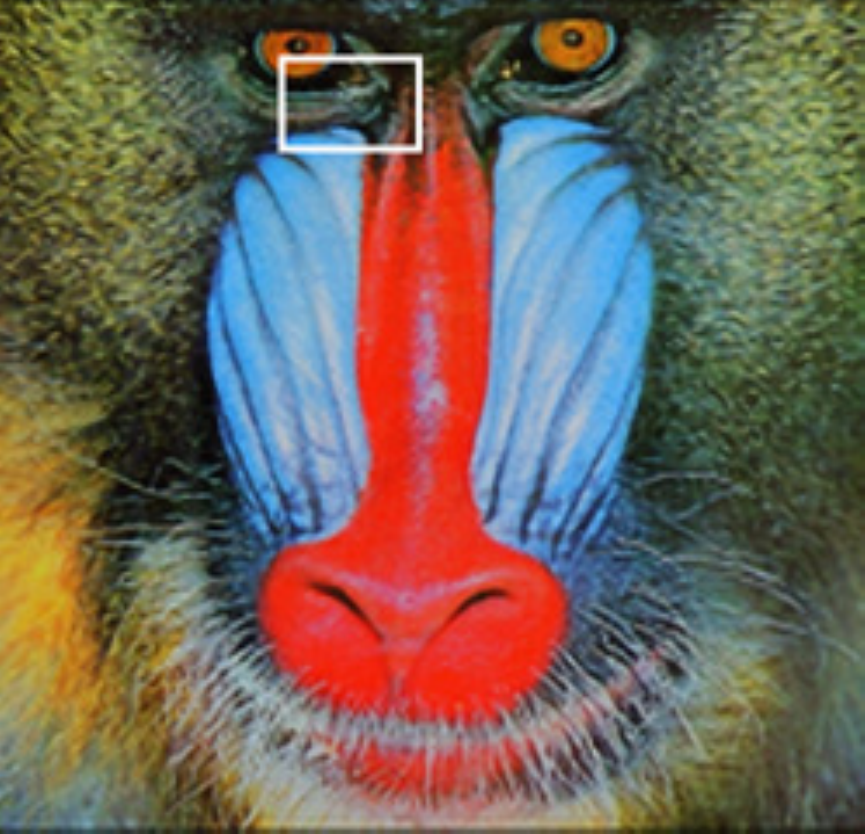}
\end{minipage}
}
  \subfloat{
\begin{minipage}[t]{0.15\textwidth}
\centering
\includegraphics[width=1\textwidth]{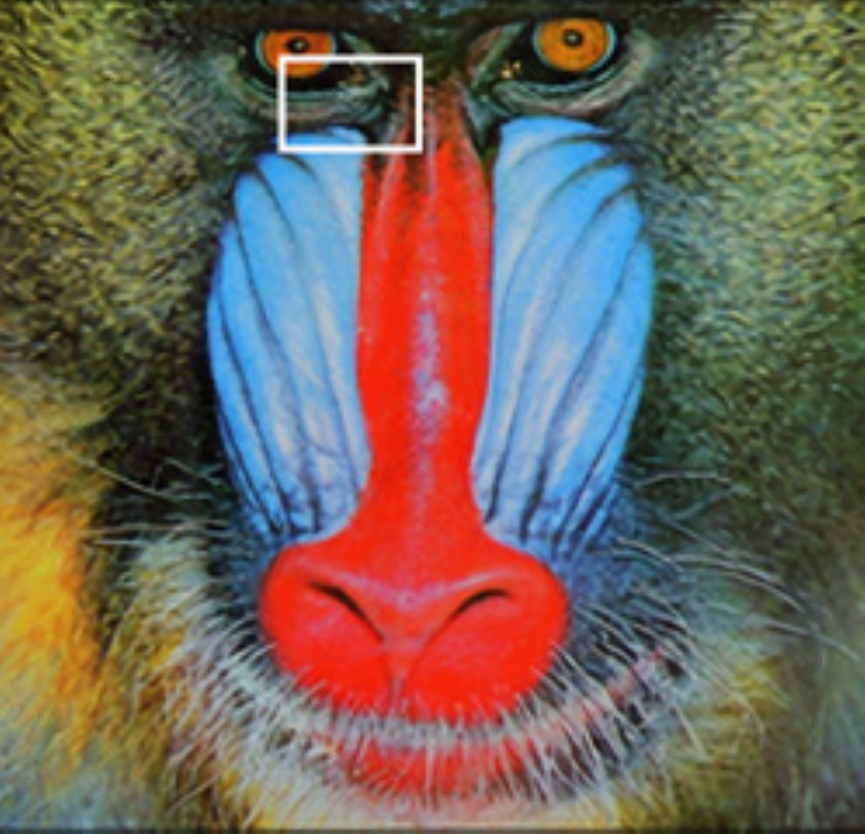}
\end{minipage}
}
  \subfloat{
\begin{minipage}[t]{0.15\textwidth}
\centering
\includegraphics[width=1\textwidth]{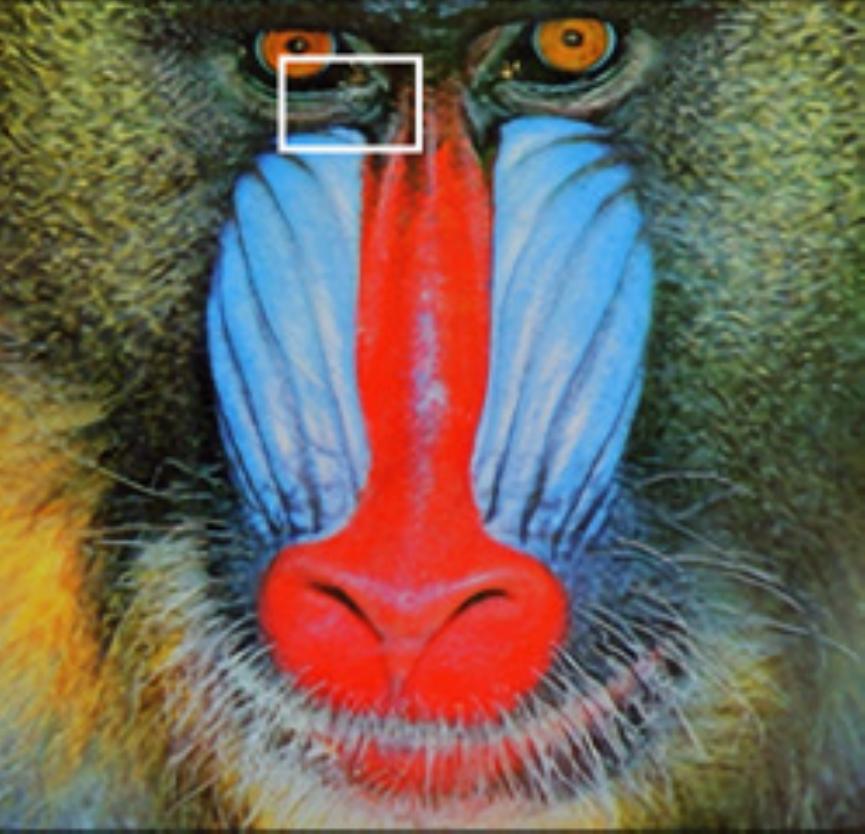}
\end{minipage}
}
  \subfloat{
\begin{minipage}[t]{0.15\textwidth}
\centering
\includegraphics[width=1\textwidth]{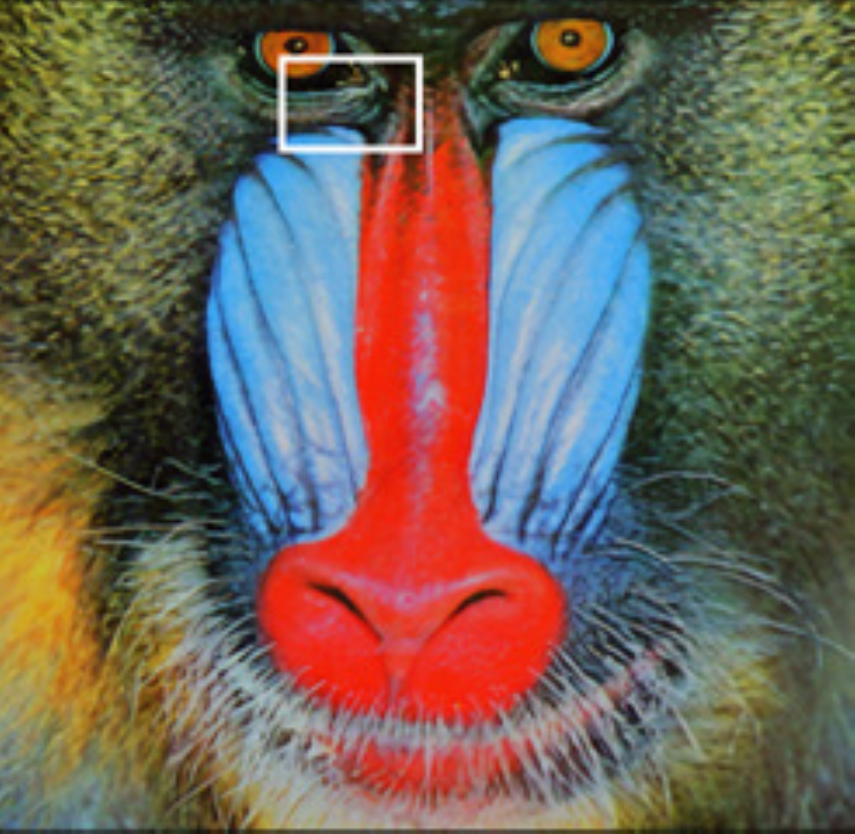}
\end{minipage}
}
  \subfloat{
\begin{minipage}[t]{0.15\textwidth}
\centering
\includegraphics[width=1\textwidth]{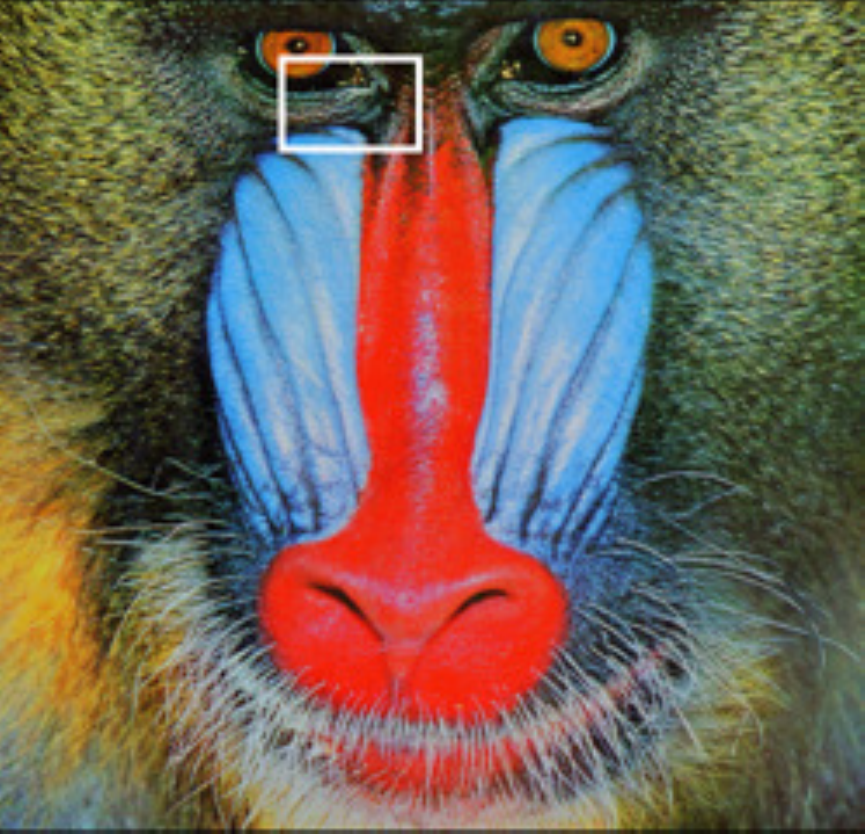}
\end{minipage}
}

  \subfloat[Bicubic]{
\begin{minipage}[t]{0.15\textwidth}
\centering
\includegraphics[width=1\textwidth]{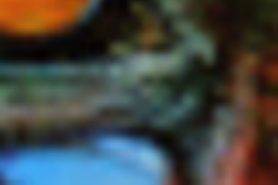}
\end{minipage}
}
  \subfloat[SRCNN]{
\begin{minipage}[t]{0.15\textwidth}
\centering
\includegraphics[width=1\textwidth]{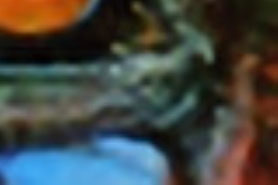}
\end{minipage}
}
  \subfloat[DRCN]{
\begin{minipage}[t]{0.15\textwidth}
\centering
\includegraphics[width=1\textwidth]{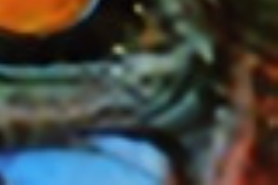}
\end{minipage}
}
  \subfloat[VDSR]{
\begin{minipage}[t]{0.15\textwidth}
\centering
\includegraphics[width=1\textwidth]{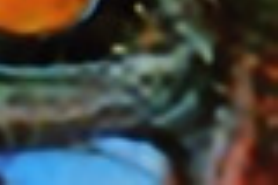}
\end{minipage}
}
  \subfloat[Proposed Method]{
\begin{minipage}[t]{0.15\textwidth}
\centering
\includegraphics[width=1\textwidth]{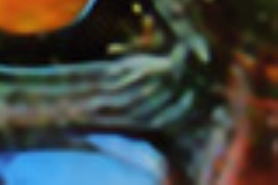}
\end{minipage}
}
  \subfloat[HR]{
\begin{minipage}[t]{0.15\textwidth}
\centering
\includegraphics[width=1\textwidth]{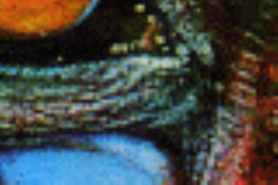}
\end{minipage}
}  
\begin{comment}
      \subfigure{
    \label{fig:50bicubic} %% label for first subfigure
    \includegraphics[width=0.14\textwidth]{small_50bicubic312.pdf}}
  \subfigure{
    \label{fig:50srcnn} %% label for first subfigure
    \includegraphics[width=0.14\textwidth]{small_50srcnn312.pdf}}
  \subfigure{
    \label{fig:50drcn} %% label for first subfigure
    \includegraphics[width=0.14\textwidth]{small_50drcn312.pdf}}
  \subfigure{
    \label{fig:50vdsr} %% label for first subfigure
    \includegraphics[width=0.14\textwidth]{small_50vdsr312.pdf}}
  \subfigure{
    \label{fig:50edsr} %% label for first subfigure
    \includegraphics[width=0.14\textwidth]{small_50edsr_mask.pdf}}
  \subfigure{
    \label{fig:50hr} %% label for first subfigure
    \includegraphics[width=0.14\textwidth]{small_50hr312.pdf}}
    \end{comment}
 %%%%%
 
    \subfloat{
\begin{minipage}[t]{0.15\textwidth}
\centering
\includegraphics[width=1\textwidth]{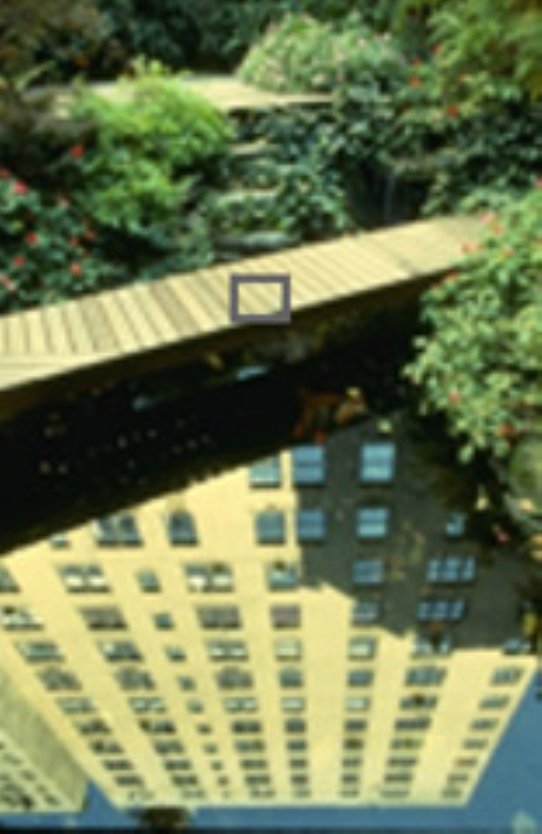}
\end{minipage}
}
  \subfloat{
\begin{minipage}[t]{0.15\textwidth}
\centering
\includegraphics[width=1\textwidth]{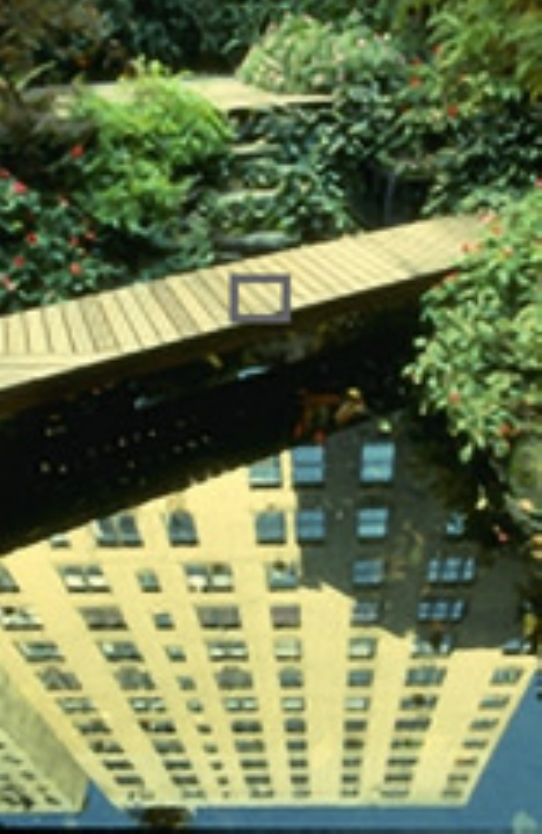}
\end{minipage}
}
  \subfloat{
\begin{minipage}[t]{0.15\textwidth}
\centering
\includegraphics[width=1\textwidth]{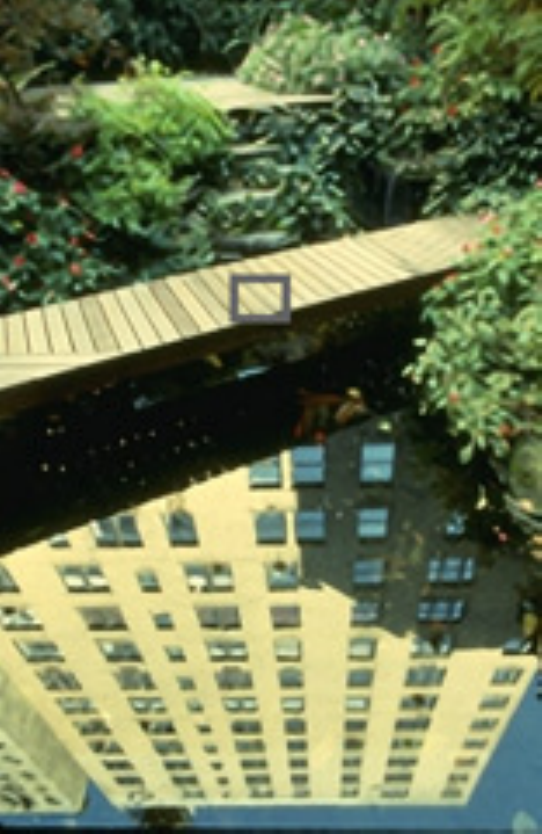}
\end{minipage}
}
  \subfloat{
\begin{minipage}[t]{0.15\textwidth}
\centering
\includegraphics[width=1\textwidth]{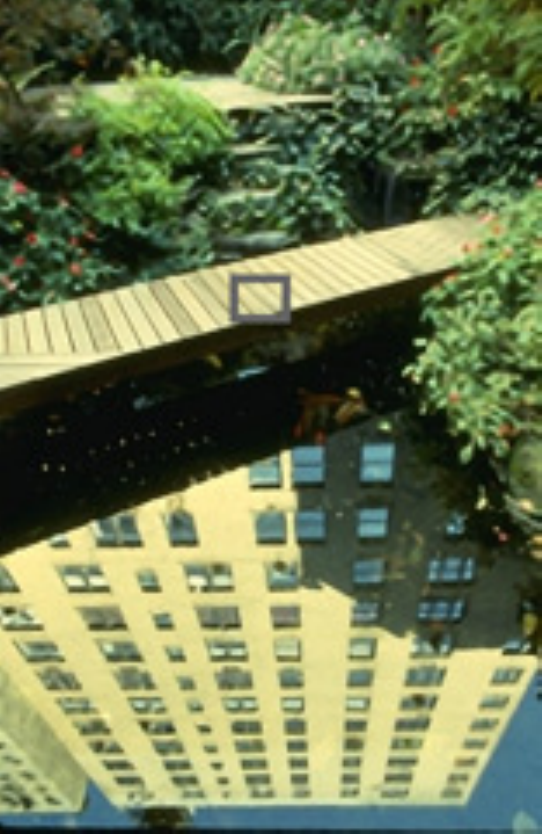}
\end{minipage}
}
  \subfloat{
\begin{minipage}[t]{0.15\textwidth}
\centering
\includegraphics[width=1\textwidth]{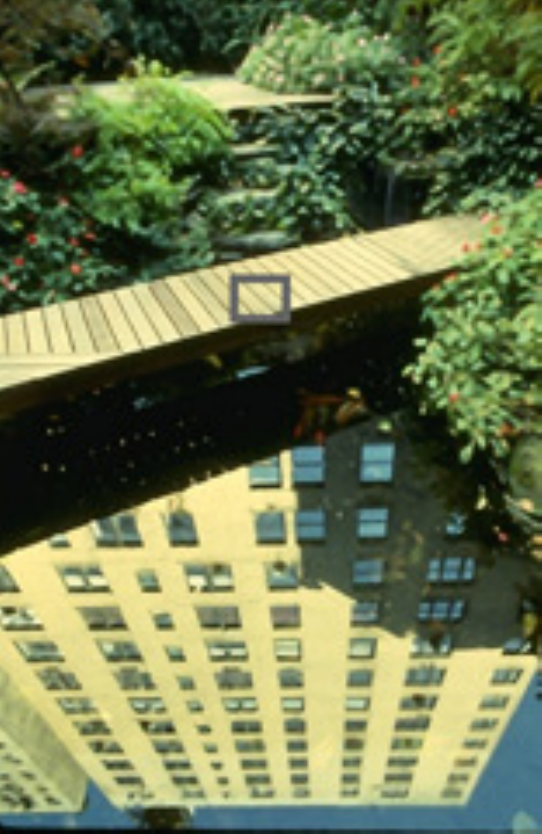}
\end{minipage}
}
  \subfloat{
\begin{minipage}[t]{0.15\textwidth}
\centering
\includegraphics[width=1\textwidth]{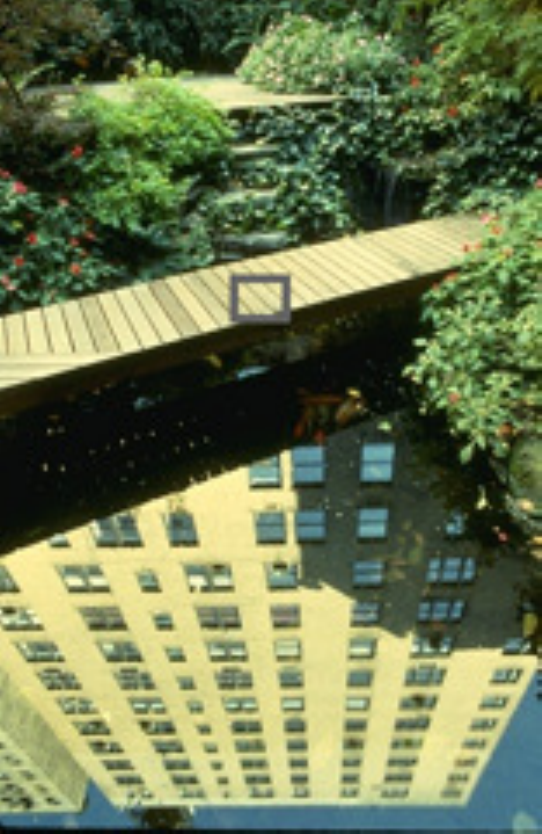}
\end{minipage}
}

   \subfloat[Bicubic]{
\begin{minipage}[t]{0.15\textwidth}
\centering
\includegraphics[width=1\textwidth]{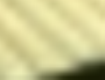}
\end{minipage}
}
  \subfloat[SRCNN]{
\begin{minipage}[t]{0.15\textwidth}
\centering
\includegraphics[width=1\textwidth]{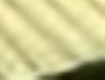}
\end{minipage}
}
  \subfloat[DRCN]{
\begin{minipage}[t]{0.15\textwidth}
\centering
\includegraphics[width=1\textwidth]{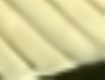}
\end{minipage}
}
  \subfloat[VDSR]{
\begin{minipage}[t]{0.15\textwidth}
\centering
\includegraphics[width=1\textwidth]{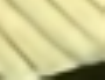}
\end{minipage}
}
  \subfloat[Proposed Method]{
\begin{minipage}[t]{0.15\textwidth}
\centering
\includegraphics[width=1\textwidth]{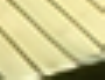}
\end{minipage}
}
  \subfloat[HR]{
\begin{minipage}[t]{0.15\textwidth}
\centering
\includegraphics[width=1\textwidth]{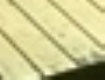}
\end{minipage}
}  

  %%%%
    \caption{Comparison of our models with other works on $\times$3 SR. The images are from Set14 \cite{set14} and BSD100 \cite{bsd300}.}
  \label{fig:the effect of different losses} 
\end{figure*}

\subsection{Comparison with  state-of-the-art methods}
In this section, we compare the results of our model with other state-of-the-art methods including SelfExSR \cite{selfex}, SRCNN \cite{learning_a_deep}, A+ \cite{A+}, VDSR \cite{accurate_image_s} and DRCN \cite{DRCN}. The performance of these methods are presented in Fig. 7. The results of SRCNN, DRCN and VDSR lack high frequency details and look blurry. On the contrary, our method can output high resolution images with high frequency details which are perceptually similar to the ground truth. 

The quantitative results by PSNR and SSIM are presented in table 1. For fair comparison, we deal with y channel only because human vision is more sensitive to details in intensity than in color. We ignore the same amount of pixels as scales from image boundary in order to eliminate the effect of zero-padding in convolution layers. MATLAB functions is used for the evaluation.
Our methods with attention producing network get the best results in the metrics of PSNR and SSIM. On average, in terms of PSNR, an improvement of 0.54 dB using the proposed method is achieved over VDSR and an improvement of 0.52 dB is achieved over DRCN. In terms of SSIM, the proposed method is also higher than other method.
On the whole, our method can achieve superior results, especially on the image with rich texture information, over many existing state-of-the-art super resolution methods.

\begin{comment}
\section{Analysis and discussion}
\end{comment}

\section{Conclusion}\label{sec:Conclusion}
We have proposed an attention-based approach to give a discrimination between texture areas and smooth areas. When the locations of the high frequency details are located, the attention mechanism works as a feature selector which enhance high frequency features and suppress noise in smooth areas. Thus, our method avoids recovering high frequency details blindly. 
We integrate the mechanism into SISR networks including SRResNet, VDSR and DRCN, and the performance of these SISR networks are all improved. Thus, the effectiveness of the attention mechanism is verified. 
As for the feature reconstruction network, we propose the DenseRes block  which provides an efficient way to combine low level features and high level features. By the cascade of multiple DenseRes blocks, our network has a large receptive field. Therefore, useful contexture information in large regions from LR images are captured to recover the high frequency details in HR images.
Our method has the best performance compared with state-of-the-art methods. In the future, we will explore applications of the attention mechanism in video super resolution to generate visually and quantitatively high quality results.

\begin{comment}
Given to the fact that SISR is a heavily ill-posed problem, we cannot produce perfect high resolution images. The multiplication operation is at the end of the network, we may go further by changing the interactive mode between the two networks of our model.
\end{comment}

\section{Acknowledgments}
This work was supported in part by the Natural Science Foundation of Jiangsu Province under Grant BK20151102, in part by the Ministry of Education Key Laboratory of Machine Perception, Peking University under Grant K-2016-03, in part by the AI lab of Southeast University, in part by the Open Project Program of the Ministry of Education Key Laboratory of Underwater Acoustic Signal Processing, Southeast University under Grant UASP1502, and in part by the Natural Science Foundation of China under Grant 61673108.

\bibliographystyle{IEEEtran}

\bibliography{IEEEabrv,sisr_conf}

\end{document}